\documentclass[runningheads]{llncs}

\usepackage{eccv}

\usepackage{eccvabbrv}

\usepackage{graphicx}
\usepackage{booktabs}

\usepackage[dvipsnames]{xcolor}
\usepackage{times}
\usepackage{epsfig}
\usepackage{graphicx}
\usepackage{amsmath}
\usepackage{amssymb}
\usepackage{floatflt}
\usepackage[singlelinecheck=false]{caption}
\usepackage[ruled,titlenumbered,vlined]{algorithm2e}
\usepackage{booktabs}
\usepackage{soul}
\usepackage{url}
\usepackage[utf8]{inputenc}
\usepackage{cuted}
\usepackage{amssymb}
\usepackage{xcolor}
\usepackage{multicol}
\usepackage{multirow}
\usepackage{makecell}
\usepackage{pifont}
\usepackage{wrapfig}
\usepackage{mathtools}
\usepackage{colortbl}
\usepackage{float}
\usepackage{tikz}
\usepackage{pgfplots}
\usepackage{pgfplotstable}
\usetikzlibrary{calc}
\pgfplotsset{compat=newest}
\usepackage{tabularx}
\newcolumntype{Y}{>{\centering\arraybackslash}X}
\definecolor{Gray}{gray}{0.85}

\definecolor{lightgreen}{RGB}{86,188,80} 
\definecolor{lightpink}{RGB}{242,224,222} 
\usepackage{colortbl}
\definecolor{LGray}{gray}{.9}
\definecolor{darkgray}{rgb}{0.66, 0.66, 0.66}

\allowdisplaybreaks
\renewcommand{\eg}{\emph{e.g.}}
\renewcommand{\ie}{i.\,e.}

\definecolor{deemph}{gray}{0.6}
\newcommand{\gc}[1]{\textcolor{deemph}{#1}}
\definecolor{forestgreen}{RGB}{34,139,34}
\newcommand{\mypar}{\\[\parskip]}
\renewcommand{\paragraph}[1]{\vspace{1mm} \noindent \textbf{{#1}}}
\makeatletter
\newcommand\Label[1]{%
  &\refstepcounter{equation}%
  (\theequation)%
  \protected@write\@auxout{}{%
    \string\newlabel{#1}{{\theequation}{\thepage}}%
  }%
  &%
}
\makeatother
\usepackage[accsupp]{axessibility}  % Improves PDF readability for those with disabilities.

\usepackage{hyperref}

\usepackage{orcidlink}

\begin{document}

\title{Event Camera Data Dense Pre-training} 

\titlerunning{Event Camera Data Dense Pre-training}

\author{Yan Yang\inst{2}
%\orcidlink{0000-1111-2222-3333} 
\and
Liyuan Pan\inst{1} 
%\orcidlink{1111-2222-3333-4444} 
\and
Liu Liu\inst{3}
%\orcidlink{2222--3333-4444-5555}
}

\authorrunning{Yan Yang et al.}

\institute{School of CSAT, Beijing Institute of Technology, Beijing, China\\Corresponding author \and
BDSI, Australian National University, Canberra, Australia \and
KooMap Dept., Huawei, Beijing, China \\
\email{Yan.Yang@anu.edu.au Liyuan.Pan@bit.edu.cn liuliu33@huawei.com}%\email{\{abc,lncs\}@uni-heidelberg.de}
}

\maketitle

\begin{abstract}
This paper introduces a self-supervised learning framework designed for pre-training neural networks tailored to dense prediction tasks using event camera data. Our approach utilizes solely event data for training.

Transferring achievements from dense RGB pre-training  directly to event camera data yields subpar performance. This is attributed to the spatial sparsity inherent in an event image (converted from event data), where many pixels do not contain information. To mitigate this sparsity issue, we encode an event image into event patch features, automatically mine contextual similarity relationships among patches, group the patch features into distinctive contexts, and enforce context-to-context similarities to learn discriminative event features.

For training our framework, we curate a synthetic event camera dataset featuring diverse scene and motion patterns.
Transfer learning performance on downstream dense prediction tasks illustrates the superiority of our method over state-of-the-art approaches. 
\keywords{Event Camera Data \and Self-supervised Learning \and Dense Prediction}
\end{abstract}

\section{Introduction}
\label{sec:intro}

An event camera asynchronously records pixel-wise brightness changes of a scene \cite{eventsurvey}. In contrast to conventional RGB cameras that capture all pixel intensities at a fixed frame rate, event cameras offer a high dynamic range and microsecond temporal resolution, and is robust to lighting changes and motion blur, showing promising applications in diverse vision tasks \cite{hmnet,ddd17,mvsec,slam}.   

\begin{wrapfigure}{r}{6cm}
    \centering
    \includegraphics[width=.98\linewidth, trim={7.2pt 15.2pt 13.2pt 6.2pt},clip]{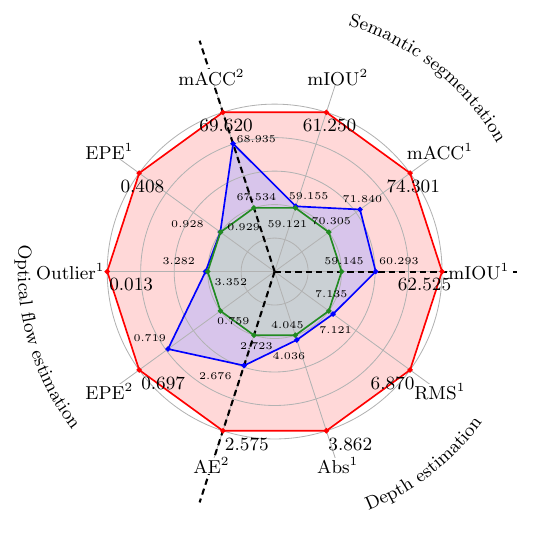}
    \vspace{-1em}
    \caption{\small \it 
    Comparison of \textcolor{red}{our} scores 
    with respect to the \textcolor{blue}{second-best} and \textcolor{forestgreen}{third-best} scores
    for semantic segmentation \cite{ddd17,evsegnet,dsec,ess}, optical flow estimation \cite{mvsec,dsec,eraft}, and depth estimation \cite{mvsec}. Superscripts besides evaluation metrics are used to differentiate benchmark datasets for a specific task.
    }
    \label{fig:overall_comaprision}
    \vspace{-.5cm}
\end{wrapfigure}

This paper addresses the task of pre-training neural networks with event camera data for dense prediction tasks, including segmentation, depth estimation, and optical flow estimation. Our self-supervised method is pre-trained solely with event camera data. One can simply transfer our pre-trained model for dense prediction tasks. Please refer to \cref{fig:overall_comaprision} for the performance comparisons. 

The direct way to pre-training is supervised training, using dense annotations for event data. However, due to the scarcity of dense annotations \cite{ddd17,mvsec,dsec}, training large-scale networks becomes challenging \cite{vit,scalinglaw}.

An alternative to supervised pre-training is self-supervised learning for event camera data \cite{eventpretraing, eclip}, which has been proposed very recently. These approaches necessitate paired RGB images and event data,  {enforcing} image-level embedding similarities between RGB images and event data. This form of RGB-guided pre-training directs networks to focus on the overall structure of events, neglecting intricate pixel-level features that are crucial for dense prediction tasks.

Next to pre-training is transferring the achievements of dense RGB pre-training \cite{densecl,esvit} to event camera data. One may first convert event camera data to an event image \cite{eventpretraing}, split the image into patches, and then learn fine-grained patch features by enforcing patch-to-patch similarities in a self-supervised learning framework. While feasible, this baseline approach is constrained as event images are sparse, containing patches with little to no information, often from the meaningless background. The sparsity diminishes the discriminativeness of an event patch, introduces background noise/bias to the patch feature learning, and makes training unstable.

Inspired by the above discriminative self-supervised approaches that learn features at the image and patch level, we show that fine-grained event features can be learned by enforcing context-level similarities among patches. Our motivation is described below. 

Given an event image, humans can recognize objects (\eg, buildings and trees) by considering multiple similar pixels. In essence, a group of event image pixels contains sufficient information to make them discriminative. Inspired by this insight, we propose to automatically mine the contextual similarity relationship among patches, group patch features into discriminative contexts, and enforce context-to-context similarities. This context-level similarity, requiring no manual annotation, not only promotes stable training but also empowers the model to achieve highly accurate dense predictions.

Our contributions are summarized as follows:
\begin{enumerate}
    \item A self-supervised framework for pre-training a backbone network for event camera dense prediction tasks. The pre-trained model can be transferred to diverse downstream dense prediction tasks;

   \item Introduction of a context-level similarity loss to address the sparsity issue of event data for learning discriminative event features;

  \item Construction of a pre-training dataset based on the TartanAir dataset \cite{tartanair}, covering diverse scenes and motion patterns to facilitate network training;

  \item State-of-the-art performance on standard event benchmark datasets for dense prediction tasks.
\end{enumerate}

\section{Related Works}
We survey recent advancements in self-supervised learning frameworks applied to RGB and event image domains. We then provide an overview of event datasets used for network pre-training and downstream task fine-tuning.

\paragraph{RGB image self-supervised learning.} Research in self-supervised learning generally falls into three categories: i) contrastive learning. Images are augmented into  multiple views for instance discrimination. By defining a matching pair (\eg, views from the same image), the similarity between them is maximized \cite{simsiam,byol}. Some works also enforce dissimilarity among non-matching pairs \cite{simclr,mocov1,mocov2,mocov3,instancediscrmination}; ii) masked image modeling. With unmasked image patches, the networks are trained to reconstruct masked ones. The reconstruction targets can be represented as intensity values of patch pixels \cite{mae,simmim}, discrete indices assigned by an image tokenizer \cite{beit,beitv2,vqgan}, or patch embeddings obtained from pre-trained vision foundation models \cite{eva,clip}; iii) self-distillation. This category can be considered as an extension of contrastive learning from instances to groups \cite{swav,dino}, and is usually combined with MIM \cite{ibot,dinov2}. The similarity between matching image pairs is optimized by minimizing a cross-entropy loss, while MIM is optionally performed. For adapting self-supervised learning frameworks to dense prediction tasks, objectives at the patch/region level are proposed to maximize the similarity between matching patches \cite{densecl,pointcl,selfpatch,esvit}. However, the spatial sparsity interferes with the patch-level objective and turns the network pre-training unstable, as most event image patches, containing little to no events, provide meaningless supervision signals. 

\paragraph{Event image self-supervised learning.} Explorations of self-supervised learning on event data remain in an early stage. Existing works \cite{eventpretraing,eclip} primarily leverage a pre-trained CLIP network \cite{clip} and paired RGB images for training, guiding the event network to have similar outputs with the RGB network (\ie, the image encoder of CLIP) in feature space. Because an event image is more similar to its paired RGB image at a high-level than at a low-level \cite{event_image_association}, these approaches concentrate on capturing the overall structures of the event image.
This explains their substantial performance improvements in object recognition tasks for event data while lagging in various dense prediction tasks. In this paper, we do not require paired RGB images and pre-trained RGB networks, and focus on pre-training a versatile network by utilizing solely event data for diverse dense prediction tasks on event datasets.

\begin{figure*}[!t]
    \centering
    \includegraphics[width=\linewidth]{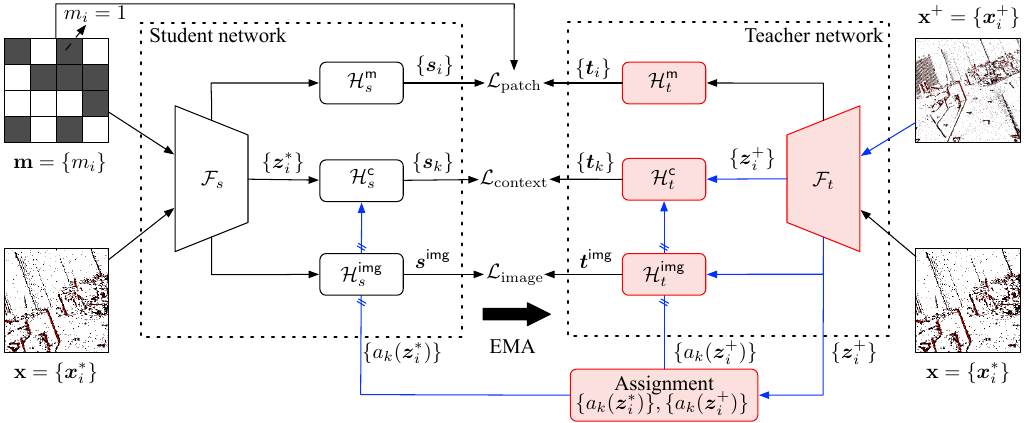}
    \caption{\small \it Overall architecture. During pre-training, our approach takes an event image $\mathbf{x}^{+}$ and its affine-transformed counterpart $\mathbf{x}^{\ast}$ as inputs, producing a pre-trained backbone network $\mathcal{F}_{s}$. A teacher network (colored by \textcolor{red}{red} boxes) and a student network are employed in the self-supervised training stage. Event images $\mathbf{x}^{+}$ and $\mathbf{x}^{\ast}$ are tiled into $N$ patches, denoted as $\mathbf{x}^{+}=\{\boldsymbol{x}^{+}_{i}\}$ and $\mathbf{x}^{\ast}=\{\boldsymbol{x}^{\ast}_{i}\}, {i=1,...,N}$. We randomly mask some patches of $\mathbf{x}^{\ast}$ given to the student, but leave $\mathbf{x}^{\ast}$ intact for the teacher. Patch-wise binary masks are represented by $\mathbf{m}=\{{m}_{i}\}$. Three similarity constraints are imposed based on output patch-wise features from the student and teacher backbones, respectively. They are: i) patch-level similarity. Patch-wise features of masked $\mathbf{x}^{\ast}$ and $\mathbf{x}^{\ast}$ are separately projected by heads $\mathcal{H}_{s}^{\mathsf{m}}$ in the student network and $\mathcal{H}_{t}^{\mathsf{m}}$ in the teacher network, obtaining embeddings $\{\boldsymbol{s}_{i}\}$ and $\{\boldsymbol{t}_{i}\}$. To reconstruct masked patch embeddings, we employ a cross-entropy loss $\mathcal{L}_{\text{patch}}$; ii) context-level similarity. Features $\{\boldsymbol{z}_{i}^{+}\}$ from the teacher network are assigned to $K$ contexts, obtaining assignments $\{a_k(\boldsymbol{z}_{i}^{+})\}$.  $a_k(\boldsymbol{z}_{i}^{+})$ denotes the membership of the feature $\boldsymbol{z}_{i}^{+}$ to $k$-th context. The assignments of student features $\{\boldsymbol{z}^{\ast}_{i}\}$ are computed by directly transferring $a_k(\boldsymbol{z}_{i}^{+})$ with an affine transformation. With the assignments $\{a_k(\boldsymbol{z}_{i}^{+})\}$ and $\{a_k(\boldsymbol{z}^{\ast}_{i})\}$, we collect and pool all features assigned to each context using heads $\mathcal{H}_{s}^{\mathsf{c}}$ and $\mathcal{H}_{t}^{\mathsf{c}}$, generating context embeddings $\{\boldsymbol{s}_{k}\}$ and $\{\boldsymbol{t}_{k}\}$. A cross-entropy loss $\mathcal{L}_{\text{context}}$ is used to learn masked context embeddings. The forward passes from $\mathbf{x}^{+}$ are colored in \textcolor{blue}{blue}, and the blocked lines mean crosslines; iii) image-level similarity. $\{\boldsymbol{z}^{\ast}_{i}\}$ and $\{\boldsymbol{z}_{i}^{+}\}$ are initially pooled separately and subsequently projected by the heads $\mathcal{H}_{s}^{\mathsf{img}}$ and $\mathcal{H}_{t}^{\mathsf{img}}$ into global image embeddings $\boldsymbol{s}^{\mathsf{img}}$ and $\boldsymbol{t}^{\mathsf{img}}$. A cross-entropy loss $\mathcal{L}_{\text{image}}$ is used to encourage image-level similarity.
    }
    \label{fig:method}
\end{figure*}

\paragraph{Event datasets.} Event cameras are bio-inspired sensors that pixel-wisely record spatial location, time, and polarity of brightness changes in a scene as an event sequence. One of the largest-scale event datasets covering diverse scenes is the N-ImageNet dataset \cite{nimagnet}. It is built by moving an event camera to observe RGB images (from the ImageNet-1K dataset \cite{imagenet}) rendered by a monitor, and inherits scene diversity from the ImageNet-1K dataset. Existing event image self-supervised learning frameworks favor leveraging the N-ImageNet dataset for pre-training, enabling transfer learning for tasks such as object recognition \cite{nimagnet,ncars,ncaltech,CIFAR-10-DVS}, depth estimation \cite{mvsec}, semantic segmentation \cite{ddd17,dsec}, and optical flow estimations \cite{mvsec,dsec}. This paper focuses on pre-training a network for the three dense prediction tasks. Moreover, considering the limited motion patterns in the N-ImageNet dataset \cite{nimagnet}, which are square, vertical, and horizontal, we curate a synthetic event dataset containing diverse motion patterns and scenes for pre-training.

\section{Method}
We present our self-supervised method in this section. Our network is trained end-to-end, and the overall architecture is shown in \cref{fig:method}.

\paragraph{Overall architecture.} We aim to learn discriminative features from event data for dense prediction tasks, such as optical flow estimation. Sharing similarities with the learning process of DINOv2 \cite{dinov2}, we convert raw events to an image \cite{voxelgrid}, and construct two event images $\mathbf{x}^{+}$ and its augmentation $\mathbf{x}^{\ast}$. The two images are then fed into teacher and student networks to learn features, followed by enforcing similarities between the features of $\mathbf{x}^{+}$ and $\mathbf{x}^{\ast}$.
We enforce three {types of} feature similarities: i) patch-level similarity; ii) context-level similarity; iii)  image-level similarity. Details of our components are provided below.

\paragraph{Event image augmentations.} We perform a 2D affine transformation on $\mathbf{x}^{+}$, followed by GaussianBlur and ColorJitter \cite{eventpretraing}, to create a distorted event image $\mathbf{x}^{\ast}$. We tile each image into $N$ patches, \ie, $\mathbf{x}^{+}=\{\boldsymbol{x}^{+}_{i}\}$ and $\mathbf{x}^{\ast}=\{\boldsymbol{x}^{\ast}_{i}\}, {i=1,...,N}$. The linearity of the affine transformation establishes pixel correspondences between $\mathbf{x}^{+}$ and $\mathbf{x}^{\ast}$. For each pixel in $\mathbf{x}^{\ast}$, we can find its corresponding pixel in $\mathbf{x}^{+}$, enabling context-level feature learning.

Image patches $\{\boldsymbol{x}^{+}_{i}\}$ and $\{\boldsymbol{x}^{\ast}_{i}\}$ are fed to the teacher and student networks for feature extraction.
In the training stage, the student network is optimized by gradient descent. To avoid model collapse, the teacher network is kept as a momentum of the student network, and its parameters are updated with an exponential moving average (EMA) \cite{mocov1}.

\paragraph{Patch-level similarity.}  
We randomly mask some patches of $\mathbf{x}^{\ast}$ given to the student, 
but leave $\mathbf{x}^{\ast}$ intact for the teacher. The goal is to reconstruct masked patch embeddings, utilizing a cross-entropy loss between the patch features of both networks on each masked patch. This objective, introduced by \cite{dinov2}, is briefly summarized below.
 
A patch-level binary mask $\mathbf{m}=\{{m}_{i}\}, i=1,...,N$ is randomly sampled. For $\boldsymbol{x}^{\ast}_{i}$, it is masked and replaced by a $\tt{[MASK]}$ token if ${m}_{i}=1$. 
The unmasked patches and $\tt{[MASK]}$ tokens are fed to the student network $\mathcal{F}_{\mathsf{s}}$ to extract features, and a feature projection head $\mathcal{H}_{s}^{\mathsf{m}}$ is employed to obtain patch embeddings $\{\boldsymbol{s}_{i}\} = \mathcal{H}_{s}^{\mathsf{m}}(\mathcal{F}_{s}(\mathbf{x}^{\ast}, \mathbf{m}))$.

Without masking, patches $\{\boldsymbol{x}^{\ast}_{i}\}$ are fed to the teacher network $\mathcal{F}_{\mathsf{t}}$ to extract features, followed by a feature projection head $\mathcal{H}_{t}$ to extract patch embeddings $\{\boldsymbol{t}_{i}\} =\mathcal{H}_{t}(\mathcal{F}_{t}(\mathbf{x}^{\ast}))$. The patch-level similarity objective is \\
\begin{minipage}{0.45\linewidth}
\begin{equation}
    \mathcal{L}_{\text{patch}} = \frac{1}{\lVert \mathbf{m} \rVert} \sum_{\substack{i=1 \\ {m}_{i}=1}}^N  
    \text{CE}\left(\boldsymbol{t}_{i}, \boldsymbol{s}_{i} \right) \ , 
    \label{eq:patch} 
\end{equation}
\end{minipage} 
\begin{minipage}{0.5\linewidth}
\vspace{-3mm}
\begin{equation}
    \text{CE}(\boldsymbol{t}, \boldsymbol{s}) = - \left\langle \mathcal{P}\left(\boldsymbol{t} \right), \log \mathcal{P} \left(\boldsymbol{s} \right) \right\rangle \ , 
    \end{equation}
\end{minipage}\\
where $\lVert \cdot \rVert$ is the L1 norm that computes the number of masked patches. $\text{CE}(\cdot,\cdot)$ is the cross-entropy loss. $\mathcal{P}(\cdot)$ is the Softmax function that normalizes the patch embedding to a distribution. $\langle\cdot,\cdot\rangle$ is the dot product.

\begin{wrapfigure}{r}{6cm}
    \centering
    
    \includegraphics[width=.88\linewidth]{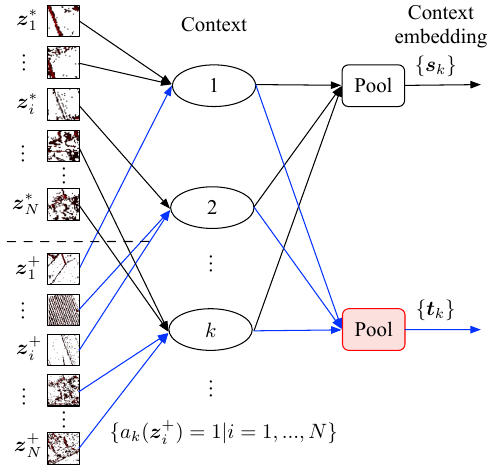}
    \caption{\small \it Context assignment and aggregation. Given patch features $\{\boldsymbol{z}^{\ast}_{i}\}$ and $\{\boldsymbol{z}_{i}^{+}\}$, we perform K-means clustering to mine $K$ contexts, and obtain the patch-to-context assignments $\{a_k(\boldsymbol{z}^{\ast}_{i})\}$ and $\{a_k(\boldsymbol{z}_{i}^{+})\}$, respectively. For the $k$-th context, $\{\boldsymbol{z}^{+}_{i}\}$ assigned to it $\{a_k(\boldsymbol{z}^{+}_{i})=1|i=1,...,N\}$ are pooled into a context embeddings $\boldsymbol{t}_k$. Similarly, $\{\boldsymbol{z}_{i}^{\ast}\}$ are pooled into context embeddings $\{\boldsymbol{s}_k\}$. The \textcolor{red}{red} box and \textcolor{blue}{blue} lines denote components of our teacher network and forward passes of $\{\boldsymbol{z}_{i}^{+}\}$, respectively.   
    }
    \label{fig:patchset}
    \vspace{-1.25cm}
\end{wrapfigure}

\paragraph{Context-level similarity.} Reconstructing each masked patch embedding independently is prone to generating noisy embeddings. This is due to the sparsity of an event image. An event patch contains little information, and many patches are from a meaningless background (see \cref{fig:cluster}). To overcome the limitations of independently reconstructing masked patch embeddings, we propose to mine contextual relationships among patch embeddings on the fly, and learn embeddings with context conditioning. We provide an overview in \cref{fig:patchset}.

Specifically, we perform K-means clustering on patch features $\{\boldsymbol{z}_{i}^{+}\} = \mathcal{F}_{t}(\mathbf{x}^{+})$ of the teacher network, generating $K$ cluster centers (\ie, contexts) and assignments $a_k(\boldsymbol{z}_{i}^{+})$. $a_k(\boldsymbol{z}_{i}^{+})$ denotes the membership of the feature $\boldsymbol{z}_{i}^{+}$ to $k$-th context, \ie, it is 1 if $\boldsymbol{z}_{i}^{+}$ is closest to $k$-th context and 0 otherwise.

For each context, features assigned to it are aggregated by an attention pooling network \cite{clip}, generating a context embedding $\boldsymbol{s}_k$. Collecting all context embeddings, we have embeddings $\{\boldsymbol{s}_k\}, k=1,...,K$, describing features  $\mathcal{F}_{t}(\mathbf{x}^{+})$ of the teacher.

For patch features $\{\boldsymbol{z}^{\ast}_{i}\} = \mathcal{F}_{s}(\mathbf{x}^{\ast}, \mathbf{m})$ of the student network, we use the same cluster centers. Due to the linearity of affine transformation, we can easily obtain the correspondence  between patches $\{\boldsymbol{x}^{\ast}_{i}\}$ and $\{\boldsymbol{x}^{+}_{i}\}$, and directly transfer the assignments $\{a_k(\boldsymbol{z}_{i}^{+})\}$ to get $\{a_k(\boldsymbol{z}^{\ast}_{i})\}$. Given assignments $a_k(\boldsymbol{z}^{\ast}_{i})$, we follow the same pipeline to aggregate features $\{\boldsymbol{z}^{\ast}_{i}\}$ into context embeddings $\{\boldsymbol{t}_k\}, k=1,...,K$. \mypar
\indent {By using adaptively mined contexts, such as roads and buildings, as proxies,}
we overcome the sparsity limitation of enforcing event patch-level similarity. In essence, we aim to enforce the similarity between a group of patches belonging to the same context. The context-level similarity loss $\mathcal{L}_{\text{context}}$ is defined below \\
\begin{minipage}{0.45\textwidth}
\begin{equation}
    \mathcal{L}_{\text{context}} = \frac{1}{K} \sum_{k=1}^K \text{CE}(\boldsymbol{t}_{k}, \boldsymbol{s}_{k}) \ .
\end{equation} 
\end{minipage}\\

\paragraph{Image-level similarity.} We aim to reconstruct masked image embedding of $\mathbf{x}^{\ast}$, by adding a cross-entropy loss between the image features of student and teacher networks on $\mathbf{x}^{\ast}$ and $\mathbf{x}^{+}$.

Patch features $\{\boldsymbol{z}^{\ast}_{i}\}$ and $\{\boldsymbol{z}_{i}^{+}\}$ from the student and teacher network $\mathcal{F}_{\mathsf{s}}$ and $\mathcal{F}_{\mathsf{t}}$ are pooled and fed to feature projection heads $\mathcal{H}_{s}^{\mathsf{img}}$ and $\mathcal{H}_{t}^{\mathsf{img}}$, generating image-level feature embeddings $\boldsymbol{s}^{\mathsf{img}}$ and $ \boldsymbol{t}^{\mathsf{img}}$, respectively. The image-level similarity objective is
\begin{equation}
    \mathcal{L}_{\text{image}} = \text{CE}\left(\boldsymbol{t}^{\mathsf{img}}, \boldsymbol{s}^{\mathsf{img}} \right) \ .
\end{equation}

\paragraph{Pre-training objective.} Our network is trained end-to-end. By using $\lambda_{1}$ and $\lambda_{2}$ hyper-parameters for balancing losses, we optimize the following objective,
\begin{equation}
    \mathcal{L}_{\text{total}} =  \mathcal{L}_{\text{patch}} + \lambda_{1} \mathcal{L}_{\text{context}} + 
    \lambda_{2} \mathcal{L}_{\text{image}} \ .
\end{equation}

\begin{table*}[!t]
    \centering
    \caption{\small \it Comparison of semantic segmentation accuracies on the DDD17 \cite{ddd17,evsegnet} and DSEC datasets \cite{dsec,ess}. Mean interaction over union (mIoU (\%)) and mean class accuracy (mACC (\%)) are used as evaluation metrics.  `\#Param', `Pre. Dataset', and `Pre. Epo.' respectively denote the number of backbone parameters, pre-training dataset, and pre-training epoch. 
    }
    \small
    \vspace{-.5em}
    \setlength{\tabcolsep}{3pt}
    \begin{tabularx}{\linewidth}{lcccccccYY}
        \toprule
         \multicolumn{1}{l}{\multirow{2}{*}{Method}} & \multirow{2}{*}{Backbone} & \multirow{2}{*}{\makecell[c]{\#Param}} & \multirow{2}{*}{\makecell[c]{Pre.\\Dataset}} &  \multirow{2}{*}{\makecell[c]{Pre.\\Epo.}}  & \multicolumn{2}{c}{DDD17} && \multicolumn{2}{c}{DSEC} \\
         \cmidrule{6-7} \cmidrule{9-10}
         &&&&& mIOU$\uparrow$ & mACC$\uparrow$  && mIOU$\uparrow$  & mACC$\uparrow$  \\
         \midrule
        \multicolumn{4}{l}{\textit{\makecell[l]{\gc{The best performance in the literature}}}} \\
        \gc{ESS} \cite{ess} & \gc{-} & \gc{-} & \gc{-} & \gc{-} & \gc{61.370} & \gc{70.874} && \gc{53.295} & \gc{62.942}  \\
        \midrule
        \multicolumn{4}{l}{\textit{\makecell[l]{\gc{Self-supervised ResNets.}}}} \\
        SimCLR \cite{simclr} & ResNet50 &23M&ImageNet-1K&100& 57.218 & 69.154 && 59.062 & 66.807 \\
        MoCo-v2 \cite{mocov2} & ResNet50  &23M&ImageNet-1K&200& 58.284 & 65.563  && 59.090 & 66.900 \\
        DenseCL \cite{densecl} & ResNet50 &23M&ImageNet-1K&200& 57.969 & 71.840 &&59.121 & 68.935\\
        ECDP \cite{eventpretraing} & ResNet50 &23M&N-ImageNet&300& 59.145 & 70.176 && {59.155} & 67.534 \\
        \rowcolor{LGray} 
        Ours & ResNet50 & 23M &  E-TartanAir & 300 & \textbf{62.912} & \textbf{74.015} && 60.641 & 69.502 \\
        \midrule 
        \multicolumn{4}{l}{\textit{\makecell[l]{\gc{Self-supervised Transformers.}}}} \\
        MoCo-v3 \cite{mocov3} & ViT-S/16 &21M&ImageNet-1K&300& 53.654 & 68.122 && 49.211 & 57.133\\
        BeiT \cite{beit} & ViT-B/16 &86M&ImageNet-1K&800& 52.391 & 61.950 && 51.899 & 59.660 \\
        IBoT \cite{ibot} & ViT-S/16 &21M&ImageNet-1K&800& 53.652 & 61.607 && 50.822 &  59.377 \\
        MAE \cite{mae} & ViT-B/16 &86M&ImageNet-1K&800& 53.758 & 64.783 && 51.958 & 59.839\\
        SelfPatch \cite{selfpatch} & ViT-S/16 &21M&ImageNet-1K&300&54.287& 62.821 &&51.475 & 59.164 \\
        DINOv2 \cite{dinov2} & ViT-S/16 &21M&LVD-142M& - &53.846&64.500&&52.165 & 59.795 \\
        CIM \cite{cim} & ViT-B/16 &86M&ImageNet-1K&300&54.013&63.926 &&51.582 & 59.628 \\
        ECDP \cite{eventpretraing} & ViT-S/16 &21M&N-ImageNet&300& 54.663 & 66.077 && 52.517 & 60.553 \\
        ESViT \cite{esvit} & Swin-T/7 &28M&ImageNet-1K&300&60.293& 70.305 &&56.517 & 63.798 \\
        \rowcolor{LGray}
        Ours & ViT-S/16 & 21M  & E-TartanAir & 300 & 55.729 & 64.771 && 56.378 & 66.000 \\
        \rowcolor{LGray}
        Ours & Swin-T/7 & 28M & E-TartanAir & 300 &{62.525} & {74.301} && \textbf{61.250} & \textbf{69.620} \\
        \bottomrule
    \end{tabularx}
    \label{tab:sem}
\end{table*}

\section{Experiments}
\noindent \textbf{Pre-training dataset.} To pre-train our network, we synthesize an E-TartanAir event camera dataset from the TartanAir dataset \cite{tartanair}. The TartanAir dataset is collected in photo-realistic simulation environments, featuring various light conditions, weather, and moving objects. It has 1037 sequences with RGB frames of $480 \times 640$ resolution. Different with N-ImageNet dataset \cite{nimagnet} that has limited motion patterns, our E-TartanAir contains diverse motion patterns and scenes. 

\paragraph{Implementation details.} We adopt ResNet50 \cite{resnet}, ViT-S/16 \cite{vit}, and Swin-T/7 architectures as our backbones. The architectures of our projection heads follow \cite{dinov2,clip}. Our model is pre-trained for 300 epochs with batch size 1024. We set $\lambda_1$ and $\lambda_2$ to 0.1 and 0.9, respectively. The number of clusters is set to 8. Our code is available at \href{https://yan98.github.io/ECDDP/}{here}.

\paragraph{Baselines.} Our method is compared against two groups of methods: i) transfer learning of self-supervised pre-training. The initial weights of state-of-the-art methods are obtained in a self-supervised manner using the ImageNet-1K \cite{imagenet}, N-ImageNet \cite{nimagnet}, or LVD-142M dataset \cite{dinov2}; ii) previous best. We compare with state-of-the-art methods specific to each downstream task, namely, semantic segmentation, flow estimation, and depth estimation. In tables, the symbols `$\downarrow$' and `$\uparrow$' indicate that a higher or lower value of a metric is preferable, respectively. The symbol `-' denotes unavailability. 

\begin{table*}[!t]
    \centering
    \caption{\it  \small
    Comparison of optical flow estimation accuracies on the MVSEC dataset \cite{mvsec}. End-point error (EPE) and outlier ratios (\%) \cite{eventpretraing} are used as evaluation metrics. Pixels with EPE above 3 and
    5\% of the ground truth optical flow magnitudes are deemed as outliers \cite{kitti}. 
    }
    \small
    \vspace{-.5em}
    \begin{tabularx}{\linewidth}{lYYYcYYcYY}
        \toprule
        \multicolumn{1}{c}{\multirow{2}{*}{Method}} & \multirow{2}{*}{Backbone} & \multicolumn{2}{c}{\textit{indoor\_flying1}} && \multicolumn{2}{c}{\textit{indoor\_flying2}} && \multicolumn{2}{c}{\textit{indoor\_flying3}} \\
        \cmidrule{3-4} \cmidrule{6-7} \cmidrule{9-10}
        & & EPE$\downarrow$ & Outlier$\downarrow$ && EPE$\downarrow$ & Outlier$\downarrow$ && EPE$\downarrow$ & Outlier$\downarrow$ \\
        \midrule
        \multicolumn{4}{l}{\textit{\makecell[l]{\gc{The best performance in the literature}}}} \\
        \gc{DCEIFlow} \cite{dceiflow} & \gc{-} & \gc{0.748} & \gc{0.597} && \gc{1.388} & \gc{8.015} && \gc{1.132} & \gc{5.294} \\
        \midrule
        \multicolumn{10}{l}{\textit{\makecell[l]{\gc{Self-supervised ResNets.}}}} \\
        SimCLR \cite{simclr} & ResNet50 & 0.646 & 0.488 && 1.445 & 9.331 && 1.188 & 5.507  \\
        MoCo-v2 \cite{mocov2} & ResNet50 & 0.612 & 0.459 && 1.359 & 8.683 && 1.130 & 5.201\\
        ECDP \cite{eventpretraing} & ResNet50 & {0.604} & 0.354 && 1.352 & 8.572  && 1.122 & 5.263\\
        DenseCL & ResNet50 & 0.634 & 0.529 && 1.349 & 7.596 && 1.130 & 5.176\\
        \rowcolor{LGray}
        Ours & ResNet50 & 0.413 & 0.055 && 0.489 & {0.041} && 0.462 & 0.007 \\ 
        \midrule
        \multicolumn{10}{l}{\textit{\makecell[l] {\gc{Self-supervised Transformers.}}}} \\
        MoCo-v3 \cite{mocov3} & ViT-S/16 & 0.648 & 0.744 && 1.361 & 8.660 && 1.119 & 5.594\\
        BeiT \cite{beit} & ViT-B/16 & 0.613 & 0.438 && 1.159 & 5.622 && 1.013 & 4.654 \\
        iBoT \cite{ibot} & ViT-S/16 & 0.630 & 0.562 && 1.259 & 6.752 && 1.038 & 4.912 \\
        MAE \cite{mae} & ViT-B/16 & 0.613 & 0.167 && 1.293 & 6.952 && 1.109 & 4.635 \\
        SelfPatch \cite{selfpatch} & ViT-S/16 &0.623&0.317&&1.337&7.894&&1.097&5.286 \\
        DINOv2 \cite{dinov2} & ViT-S/16 &0.602&0.325&&1.196&6.185&&0.990&4.333 \\
        CIM \cite{cim} & ViT-B/16 &0.625&0.491&&1.332&8.926&&1.040&4.869 \\
        ECDP  \cite{eventpretraing} & ViT-S/16 & 0.614 & {0.046} && {1.261} & {6.689}  && {1.001} & {3.111}\\
        ESViT \cite{esvit} & Swin-T/7 &0.812&1.224&&1.338&8.316&&1.078&5.185 \\
        \rowcolor{LGray}
        Ours & ViT-S/16 & 0.508 & 0.112 && 0.691 & 0.290 && 0.610 & 0.075\\ 
        \rowcolor{LGray}
        Ours & Swin-T/7 & \textbf{0.362} & \textbf{0.035} && \textbf{0.445} & \textbf{0.002} && \textbf{0.417} & \textbf{0.001}  \\
        \bottomrule
     \end{tabularx}
    \label{tab:mvsecflow}
\end{table*}
\begin{table*}[!t]
    \vspace{-1em}
    \centering
    \caption{\small \it Comparisons of optical flow estimation accuracies on the DSEC dataset \cite{dsec}. Note that IDNet, ranking first previously, maintains anonymity at the time of submission. According to the DSEC leaderboard, we present results for 1/2/3-pixel error (1/2/3-PE), end-point error (EPE), and angular error (AE). All data is sourced from the online benchmark at the time of submission.
     }
    \small
    \vspace{-.5em}
    \begin{tabularx}{\linewidth}{lYYYYY}
        \toprule
        Methods & 1PE$\downarrow$ & 2PE$\downarrow$ & 3PE$\downarrow$ & EPE$\downarrow$ & AE$\downarrow$ \\
        \midrule
        E-RAFT \cite{eraft} & 12.742 & 4.740 & 2.684 & 0.788 & 2.851\\
        MultiCM \cite{multicm} & 76.570 & 48.480 & 30.855 & 3.472 & 13.983 \\
        E-Flowformer \cite{eflowformer} & 11.225 & 4.102 & 2.446 & 0.759 & 2.676\\
        TMA \cite{tma} & 10.863 & 3.972 & 2.301 & 0.743 & 2.684 \\
        OF\_EV\_SNN \cite{onevflow} & 53.671 & 20.238 & 10.308 & 1.707 & 6.338 \\
        IDNet \cite{idnet} & 10.069 & 3.497 & 2.036 & 0.719 & 2.723 \\
        \rowcolor{LGray}
        Ours (ResNet50) & 9.013 & 3.290 & 1.983 & 0.701 & 2.611\\
        \rowcolor{LGray} 
        Ours (ViT-S/16) & 9.288 & 3.339 & 2.005 & 0.714 & 2.615 \\
        \rowcolor{LGray}
        Ours (Swin-T/7) & \textbf{8.887} & \textbf{3.199} & \textbf{1.958} & \textbf{0.697} & \textbf{2.575}\\
        \bottomrule
    \end{tabularx}
    \label{tab:descflow}
\end{table*}

\subsection{Semantic Segmentation}

\noindent \textbf{Settings.} Following the setup of \cite{eventpretraing}, we evaluate on the  
DDD17 \cite{ddd17,evsegnet} and DSEC dataset \cite{dsec,ess} for semantic segmentation. The two datasets contain selected intervals of multiple event sequences, covering 6 and 11 semantic classes, respectively. 

\paragraph{Results.} \cref{tab:sem} gives the comparisons on the DDD17 and DSEC datasets. Consistently, our method surpasses the state-of-the-art methods within the backbone groups of ResNet50, ViT-S/16, and Swin-T/7, and achieves a better performance than the methods pre-trained with a larger ViT-B/16 backbone. For example, our method with a Swin-T/7 backbone achieves mIoU/mACC scores at 62.525\%/74.301\% and 61.250\%/69.620\% on the DDD17 and DSEC datasets, respectively, outperforming all other methods. Even though DINOv2 \cite{dinov2} is trained on the huge LVD-142M dataset, our method significantly outperforms it.

\begin{table*}[!t]
    \centering
    \caption{\it  \small
    Comparison of depth estimation accuracies on the MVSEC dataset \cite{mvsec}. Averaged scores across all sequences with a cutoff threshold at 30 meters are reported. Threshold accuracy ($\delta1$, $\delta2$, and $\delta3$), absolute error (Abs), root mean squared error (RMS), and root mean squared logarithmic error (RMSlog) are used as evaluation metrics. The inputs of HMNet$^{1}$ are events, and HMNet$^{2}$ additionally takes RGB frames as inputs.
    }
    \small
    \vspace{-.5em}
    \begin{tabularx}{\linewidth}{lYYYYYYYY}
        \toprule
        \multicolumn{1}{c}{Method} & Backbone  & $\delta1$$\uparrow$ & $\delta2$$\uparrow$ & $\delta3$$\uparrow$ & Abs$\downarrow$ & RMS$\downarrow$ & RMSlog$\downarrow$  \\
        \midrule
        \multicolumn{8}{l}{\textit{\makecell[l]{\gc{The best performance in the literature.}}}} \\ 
        \gc{HMNet$^{1}$} \cite{hmnet} & \gc{-} &  \gc{0.626} & \gc{0.818} & \gc{0.912} & \gc{2.882} & \gc{4.772} & \gc{0.361} \\
        \gc{HMNet$^{2}$} \cite{hmnet} & \gc{-} & \gc{0.628} & \gc{0.803} & \gc{0.905} & \gc{2.908} & \gc{4.858} & \gc{0.359} \\
        \midrule
        \multicolumn{8}{l}{\textit{\makecell[l]{\gc{Self-supervised ResNets.}}}} \\
        SimCLR \cite{simclr} &ResNet50& 0.633 & 0.822 & 0.918 & 2.886 & 4.612 & 0.351 \\
        MoCo-v2 \cite{mocov2} & ResNet50 & 0.647  & 0.827 & 0.919 & 2.817 & 4.556 & 0.346  \\
        ECDP \cite{eventpretraing} & ResNet50 & 0.651 & 0.829 & 0.921 &2.798 & 4.530 & 0.343\\
        DenseCL \cite{densecl} & ResNet50 & 0.649 & 0.826 & 0.920 & 2.813 & 4.541 & 0.344  \\
        \rowcolor{LGray}
        Ours & ResNet50 & 0.649 & \textbf{0.837} & \textbf{0.931} & 2.713 & 4.302 & \textbf{0.330} \\
        \midrule
        \multicolumn{8}{l}{\textit{\makecell[l] {\gc{Self-supervised Transformers.}}}} \\
        MoCo-v3 \cite{mocov3} & ViT-S/16 & 0.630 & 0.814 & 0.909 & 3.043 & 4.817 & 0.362 \\
        BeiT \cite{beit} & ViT-B/16 & 0.622 & 0.805 & 0.903 & 3.147 & 4.965 & 0.372 \\
        iBoT \cite{ibot} & ViT-S/16 & 0.623 & 0.816 & 0.912 & 2.998 & 4.736 & 0.360 \\
        MAE \cite{mae} & ViT-B/16 & 0.612 & 0.802 & 0.900 & 3.214 & 5.075 & 0.377 \\
        SelfPatch \cite{selfpatch} & ViT-S/16 & 0.605 & 0.801 & 0.900 & 3.435 & 5.067 & 0.380 \\
        DINOv2 \cite{dinov2} & ViT-S/16 &  0.612 & 0.805 & 0.903 & 3.181 & 5.030 & 0.375 \\
        CIM \cite{cim} & ViT-B/16 & 0.625 & 0.808 & 0.904 & 3.108 & 4.906 & 0.370 \\
        ECDP  \cite{eventpretraing} & ViT-S/16 & 0.614 & 0.802 & 0.899 & 3.228 & 5.104 & 0.378 \\
        ESViT \cite{esvit} & Swin-T/7 &  0.644 & 0.829 & 0.923 & 2.796 & 4.482 & 0.342 \\
        \rowcolor{LGray}
        Ours & ViT-S/16 & 0.649 & 0.827 & 0.920 & 2.815 & 4.476 & 0.343 \\
        \rowcolor{LGray}
        Ours & Swin-T/7 & \textbf{0.658} & \textbf{0.837} & {0.928} & \textbf{2.658} & \textbf{4.257} & \textbf{0.330} \\
        \bottomrule
     \end{tabularx}
    \label{tab:mvsecdepth}
\end{table*}

\begin{table*}[!t]
    \centering
    \caption{\it  \small
    Comparison of depth estimation accuracies on the MVSEC dataset \cite{mvsec}. Averaged scores across all sequences are reported. Threshold accuracy ($\delta1$, $\delta2$, and $\delta3$), absolute error (Abs), root mean squared error (RMS), and root mean squared logarithmic error (RMSlog) are used as evaluation metrics. The inputs of HMNet$^{1}$ are events, and HMNet$^{2}$ additionally takes RGB frames as inputs.
    }
    \small
    \vspace{-.5em}
    \begin{tabularx}{\linewidth}{lYYYYYYYY}
        \toprule
        \multicolumn{1}{c}{Method} & Backbone  & $\delta1$$\uparrow$ & $\delta2$$\uparrow$ & $\delta3$$\uparrow$ & Abs$\downarrow$ & RMS$\downarrow$ & RMSlog$\downarrow$  \\
        \midrule
        \multicolumn{8}{l}{\textit{\makecell[l]{\gc{The best performance in the literature.}}}} \\ 
        \gc{HMNet$^{1}$} \cite{hmnet} & \gc{-} &   \gc{0.588} & \gc{0.784} & \gc{0.889} & \gc{4.171} & \gc{7.534} &\gc{0.397}\\
        \gc{HMNet$^{2}$} \cite{hmnet} & \gc{-} & \gc{0.582} & \gc{0.754} & \gc{0.860} &\gc{4.614}&\gc{8.602}&\gc{0.430} \\
        \midrule
        \multicolumn{8}{l}{\textit{\makecell[l]{\gc{Self-supervised ResNets.}}}} \\
        SimCLR \cite{simclr} &ResNet50& 0.594 & 0.789 & 0.897 & 4.176 & 7.343 & 0.386\\
        MoCo-v2 \cite{mocov2} & ResNet50 & 0.609 & 0.797 & 0.901 & 4.045 & 7.135 & 0.377 \\
        ECDP \cite{eventpretraing} & ResNet50 & {0.611} & {0.797} & 0.901 & 4.061 & 7.197 & 0.377 \\
        DenseCL \cite{densecl} & ResNet50 & 0.610 & 0.798 & 0.903 & 4.036 & {7.121} & {0.375} \\
        \rowcolor{LGray}
        Ours & ResNet50 & 0.612 & \textbf{0.809} & \textbf{0.915} & 3.889 & \textbf{6.805} & \textbf{0.359}\\
        \midrule
        \multicolumn{8}{l}{\textit{\makecell[l] {\gc{Self-supervised Transformers.}}}} \\
        MoCo-v3 \cite{mocov3} & ViT-S/16 & 0.590 & 0.782 & 0.891 & 4.313 & 7.466 & 0.394\\
        BeiT \cite{beit} & ViT-B/16 &  0.584 & 0.775 & 0.886 & 4.398 & 7.562 & 0.402 \\
        iBoT \cite{ibot} & ViT-S/16 & 0.583 & 0.782 & 0.892 & 4.309 & 7.521 & 0.394\\
        MAE \cite{mae} & ViT-B/16 &  0.575 & 0.772 & 0.884 & 4.449 & 7.601 & 0.405\\
        SelfPatch \cite{selfpatch} & ViT-S/16 &  0.567 & 0.768 & 0.882 & 4.515 & 7.735 & 0.410\\
        DINOv2 \cite{dinov2} & ViT-S/16 &  0.575 & 0.774 & 0.885 & 4.449 & 7.653 & 0.406\\
        CIM \cite{cim} & ViT-B/16 & 0.585 & 0.777 & 0.888 & 4.356 & 7.495 & 0.398\\
        ECDP  \cite{eventpretraing} & ViT-S/16 &  0.576 & 0.772 & 0.883 & 4.491 & 7.680 & 0.406 \\
        ESViT \cite{esvit} & Swin-T/7 &  0.604 & 0.796 & {0.903} & 4.083 & 7.219 & 0.377\\
        \rowcolor{LGray}
        Ours & ViT-S/16 & 0.610 & 0.800 & 0.906 & 3.987 & 6.957 & 0.369 \\
        \rowcolor{LGray}
        Ours & Swin-T/7 &  \textbf{0.618} & {0.806} & {0.912} & \textbf{3.862} & {6.870} & {0.360}\\
        \bottomrule
     \end{tabularx}

    \label{tab:mvsecdepth_cutoff}
\end{table*}

\subsection{Flow Estimation}
\paragraph{Settings.} We compare our method with state-of-the-art methods on the MVSEC dataset \cite{mvsec}. End-point error (EPE) and outlier ratios (\%) are used as evaluation metrics \cite{eventpretraing,dceiflow}. {In accordance with \cite{eventpretraing},} the evaluations are performed on the ‘indoor\_flying1’, ‘indoor\_flying2’, and ‘indoor\_flying3’ sequences. 

Additionally, our method is evaluated on the DSEC-Flow benchmark\footnote{\url{https://dsec.ifi.uzh.ch/uzh/dsec-flow-optical-flow-benchmark/}}\cite{dsec, eraft}, securing the first-place position at the time of submission.

\paragraph{Results.} \cref{tab:mvsecflow} presents the comparisons on MVSEC dataset. Among the three different backbone groups, we have the most accurate flow estimation by pre-training with a Swin-T/7 backbone, and the EPE and outlier ratios on the three sequences are 0.362/0.035\%, 0.445/0.002\%, and 0.417/0.001\%, respectively, which are significantly better than all other methods.

Results for the DSEC-Flow benchmark are given in \cref{tab:descflow}. Compared with the method IDNet \cite{idnet}, previously holding the top position, our method achieves superior optical flow estimation accuracy. For example, our method with a ResNet50, ViT-S/16, and Swin-T/7 backbone respectively improves the state-of-the-art EPE/AE scores from 0.719/2.723 to 0.701/2.611, 0.714/2.615, and 0.697/2.575.

\begin{figure*}[!t]
    \centering
    \bgroup
    \setlength{\tabcolsep}{2.5pt}
    \def\arraystretch{1.5}
        \begin{tabular}{ccccccc}
        \includegraphics[width=0.155\textwidth,height = .109\textwidth]{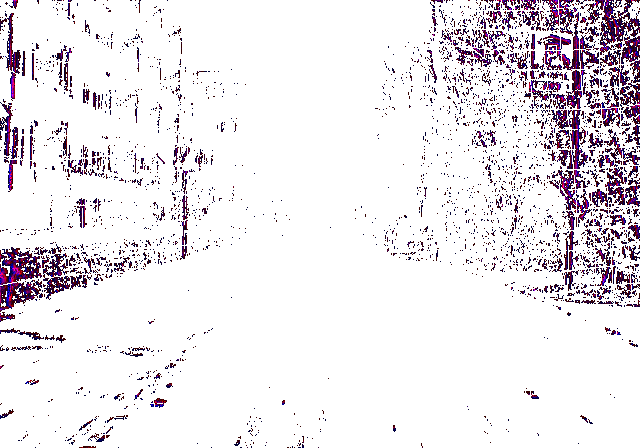} & 
        \includegraphics[width=0.155\textwidth,height = .109\textwidth]{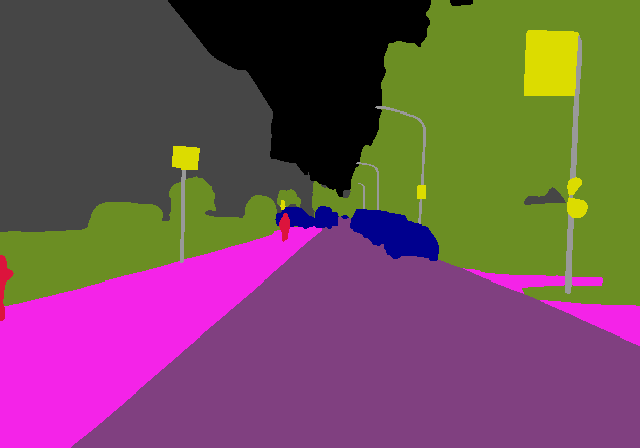} & 
        \includegraphics[width=0.155\textwidth,height = .109\textwidth]{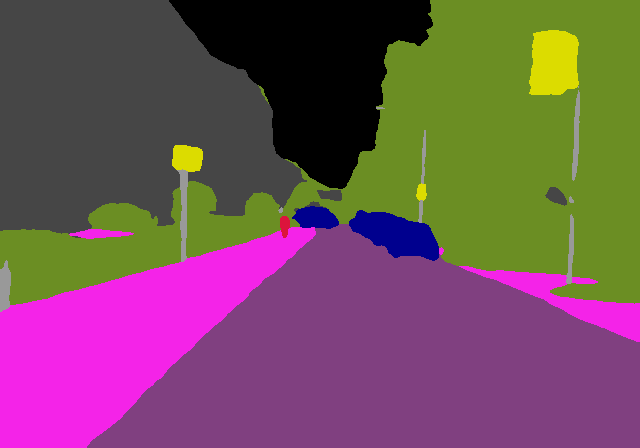} & 
        \includegraphics[width=0.155\textwidth, height = .109\textwidth, trim={1cm 0 2cm 4.9cm},clip]{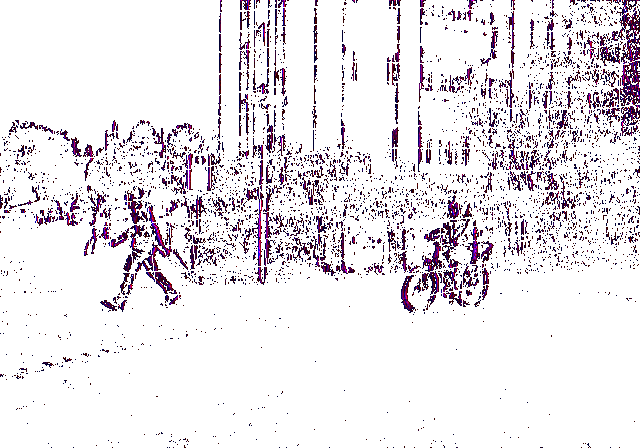} & 
        \includegraphics[width=0.155\textwidth, height = .109\textwidth, trim={1cm 0 2cm 4.9cm},clip]{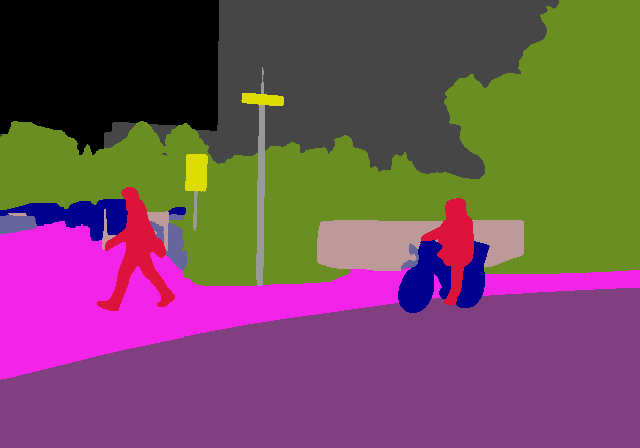} & 
        \includegraphics[width=0.155\textwidth, height = .109\textwidth, trim={1cm 0 2cm 4.9cm},clip]{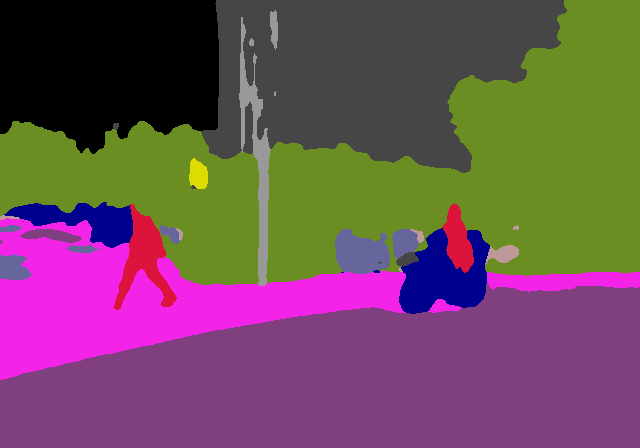}  \\
        \includegraphics[width=0.155\textwidth]{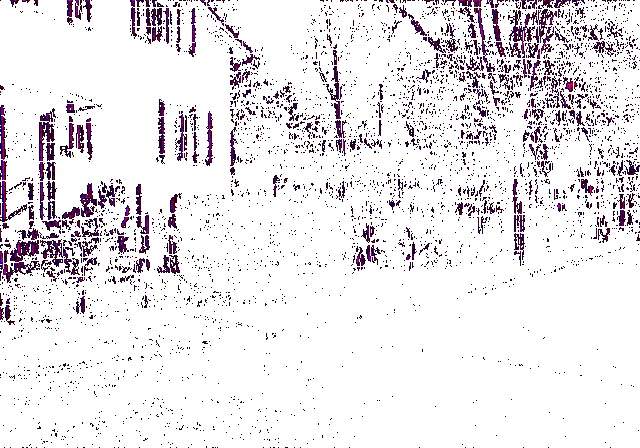} & 
        \includegraphics[width=0.155\textwidth]{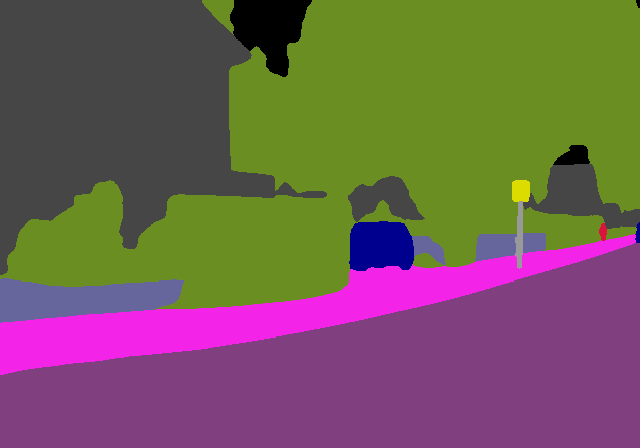} & 
        \includegraphics[width=0.155\textwidth]{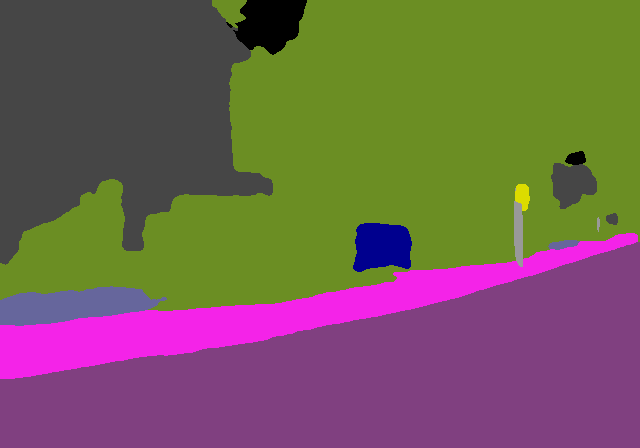} & 
        \includegraphics[width=0.155\textwidth]{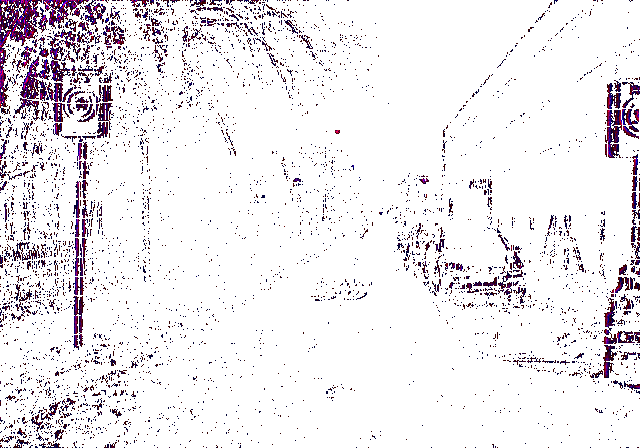} & 
        \includegraphics[width=0.155\textwidth]{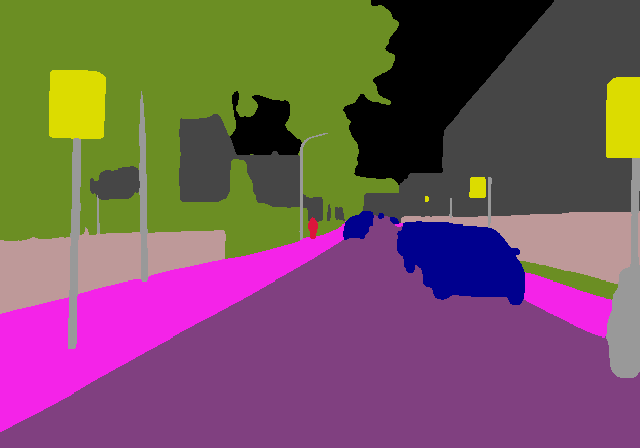} & 
        \includegraphics[width=0.155\textwidth]{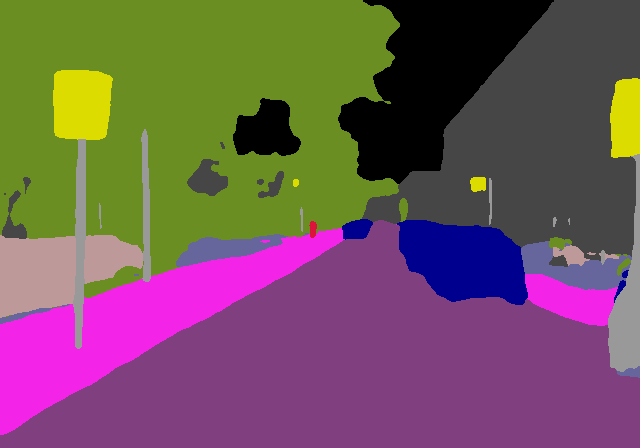}  \\
        \includegraphics[width=0.155\textwidth, height = .11625\textwidth]{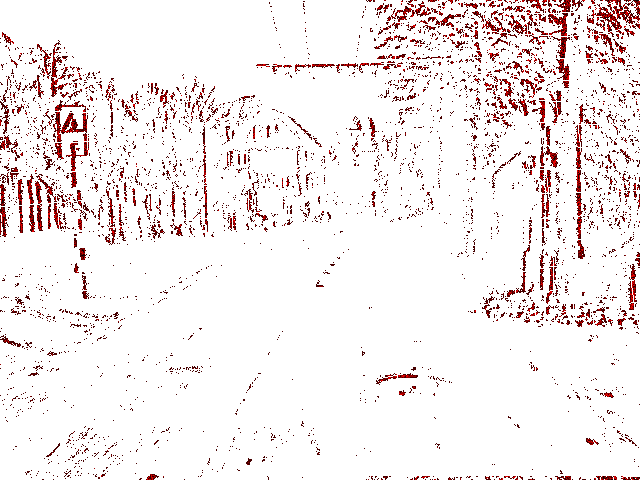} & 
        \includegraphics[width=0.155\textwidth, height = .11625\textwidth]{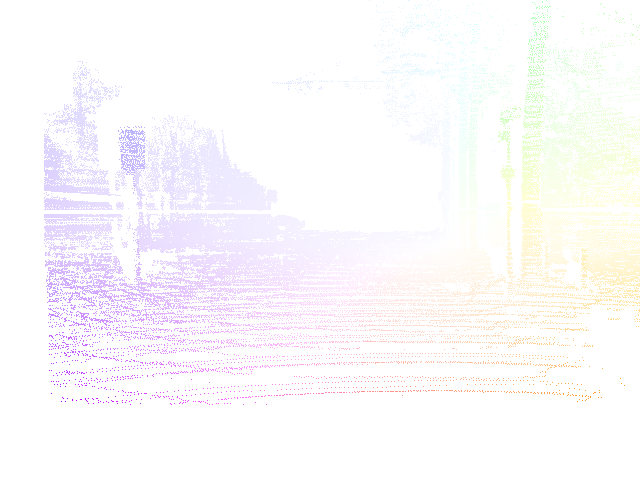} & 
        \includegraphics[width=0.155\textwidth, height = .11625\textwidth]{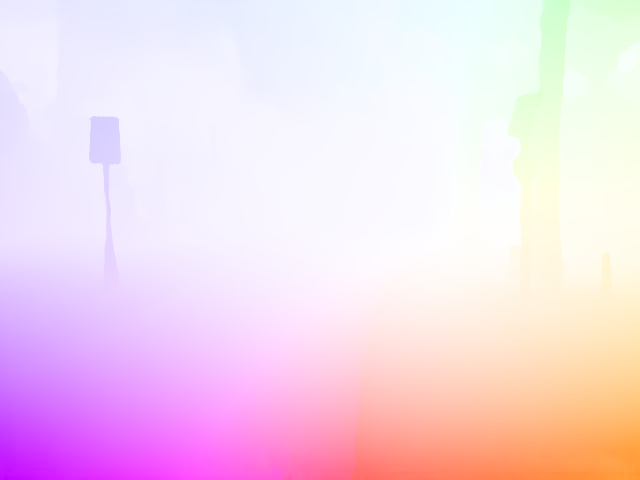} & 
        \includegraphics[width=0.155\textwidth, height = .11625\textwidth, trim={0 0.6cm 0 0},clip]{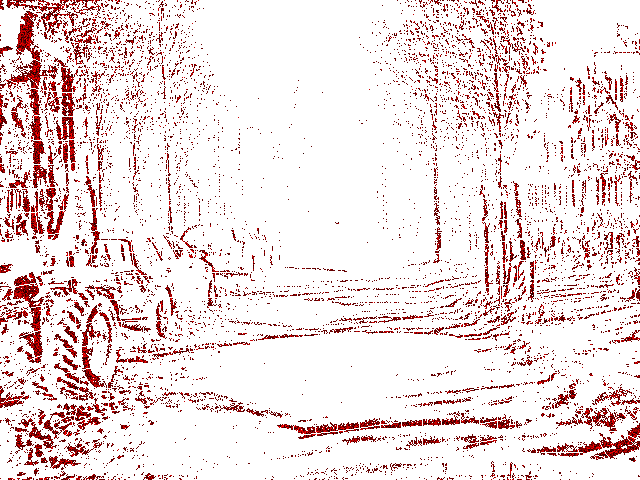} & 
        \includegraphics[width=0.155\textwidth, height = .11625\textwidth, trim={0 0.6cm 0 0},clip]{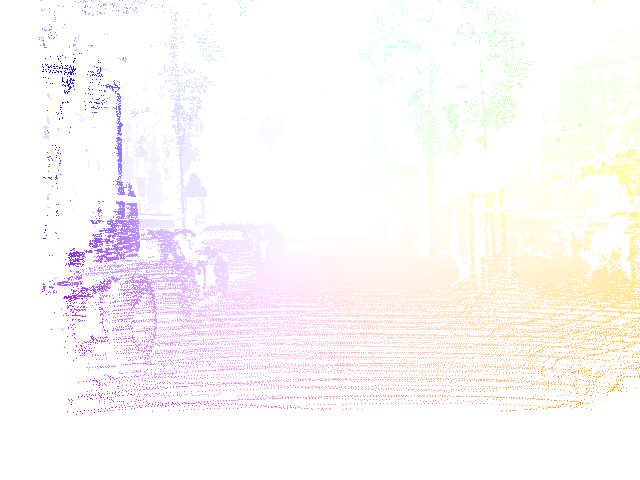} & 
        \includegraphics[width=0.155\textwidth, height = .11625\textwidth, trim={0 0.6cm 0 0},clip]{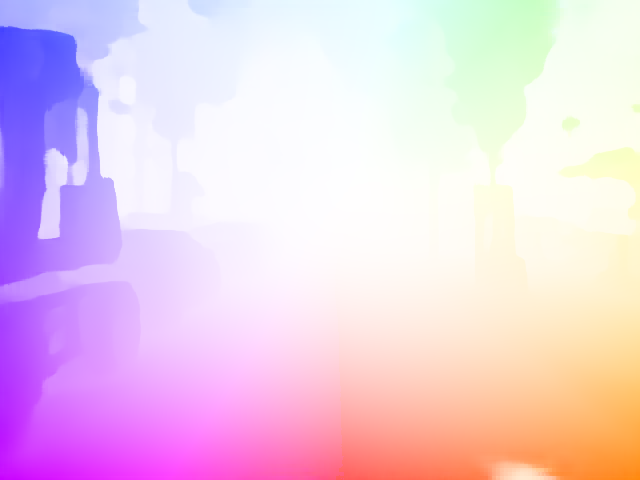} \\
        \includegraphics[width=0.155\textwidth, height = .11625\textwidth]{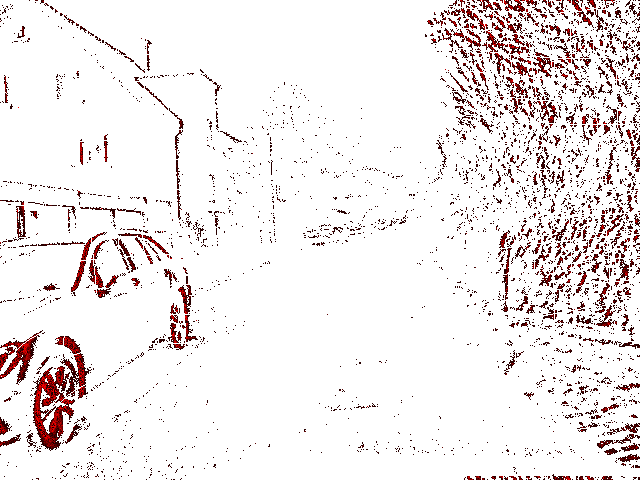} & 
        \includegraphics[width=0.155\textwidth, height = .11625\textwidth]{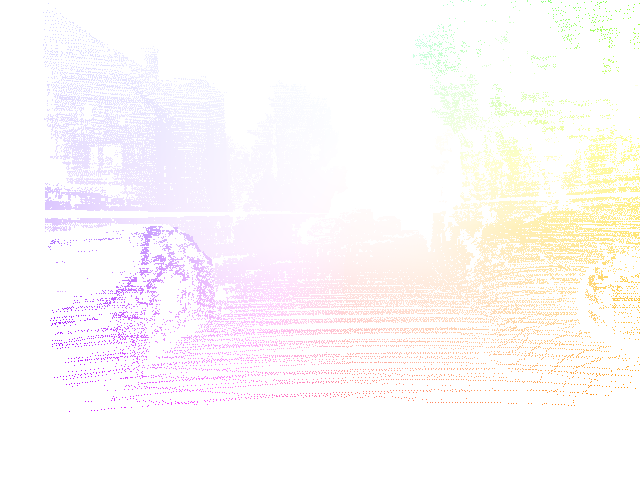} & 
        \includegraphics[width=0.155\textwidth, height = .11625\textwidth]{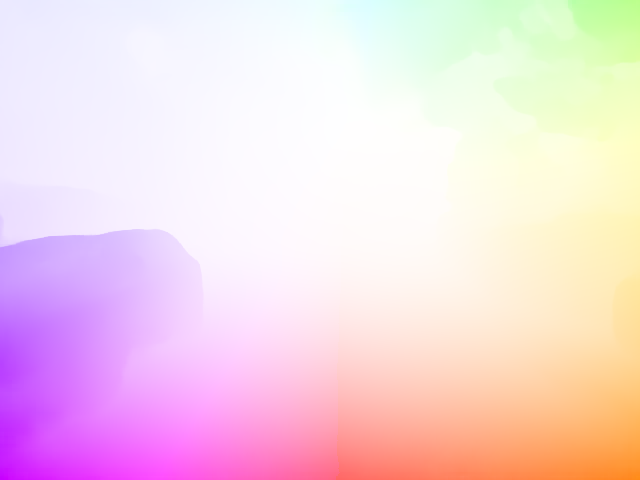} &
        \includegraphics[width=0.155\textwidth, height = .11625\textwidth]{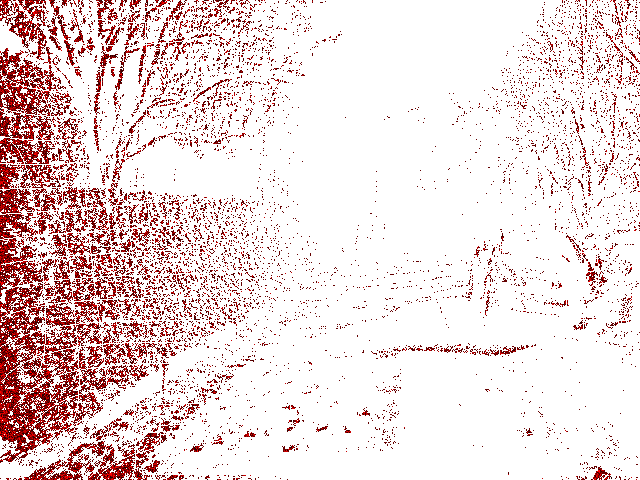} & 
        \includegraphics[width=0.155\textwidth, height = .11625\textwidth]{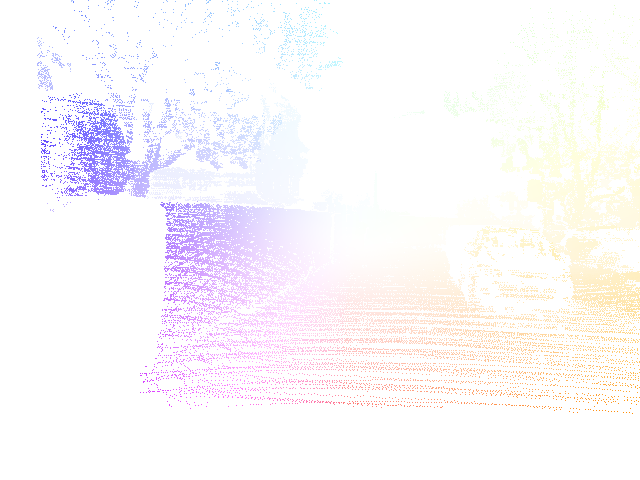} & 
        \includegraphics[width=0.155\textwidth, height = .11625\textwidth]{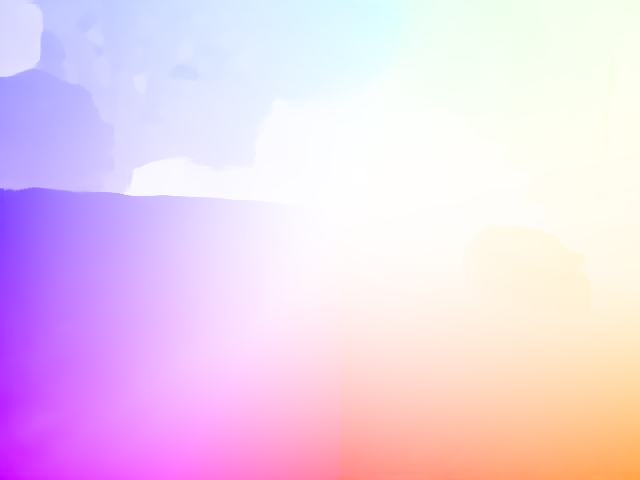} \\
        \includegraphics[width=0.155\textwidth]{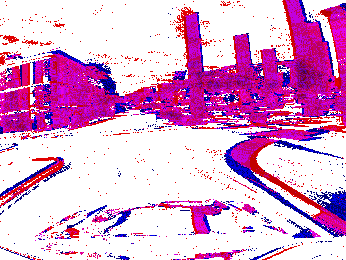} & 
        \includegraphics[width=0.155\textwidth]{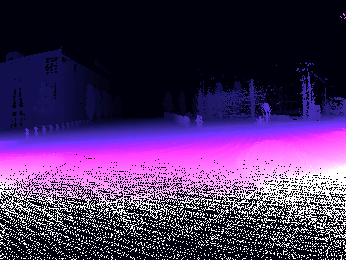} & 
        \includegraphics[width=0.155\textwidth]{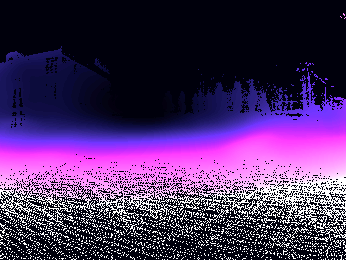} & 
        \includegraphics[width=0.155\textwidth]{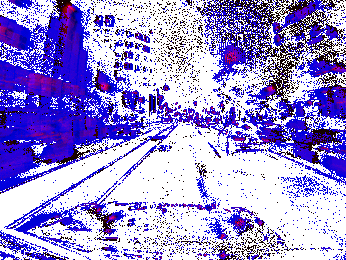} & 
        \includegraphics[width=0.155\textwidth]{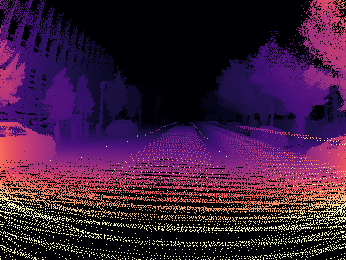} & 
        \includegraphics[width=0.155\textwidth]{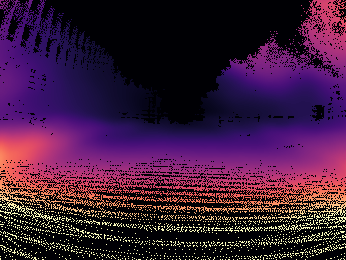} \\
        \includegraphics[width=0.155\textwidth]{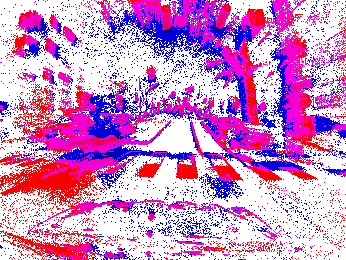} & 
        \includegraphics[width=0.155\textwidth]{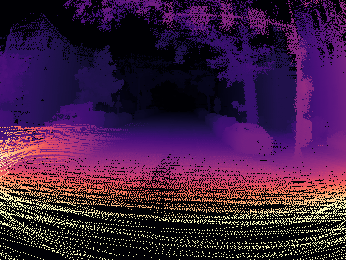} & 
        \includegraphics[width=0.155\textwidth]{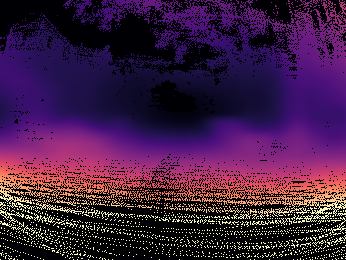} & 
        \includegraphics[width=0.155\textwidth]{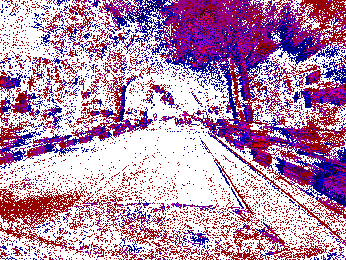} & 
        \includegraphics[width=0.155\textwidth]{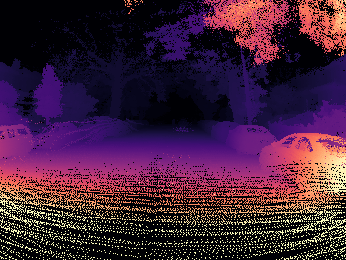} & 
        \includegraphics[width=0.155\textwidth]{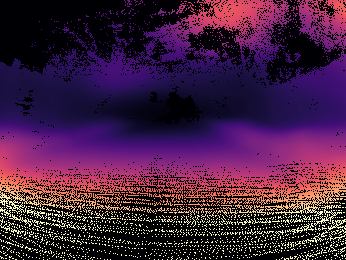} \\
       (a) & (b) & (c) &  (d) & (e) & (f) \\
    \end{tabular}
    \egroup
    \vspace{-.5em}
    \caption{\it \small Qualitative comparison examples of dense predictions, namely, semantic segmentation (1$^\text{st}$-2$^\text{nd}$ rows), optical flow estimation (3$^\text{rd}$-4$^\text{th}$ rows), and depth estimation (5$^\text{th}$-6$^\text{th}$ rows).
     (a) and (d): event images. Red and blue pixels depict positive and negative events, respectively. (b) and (e): ground-truth labels. (c) and (f): our model predictions. The brightness of depth maps in the 5$^\text{th}$ row of (b) and (c) is enhanced for visualization. 
    }
    \label{fig:seg_mainpaper}
\end{figure*}

\subsection{Depth Estimation}

\paragraph{Settings.} We evaluate the performance of our methods for depth estimation on the MVSEC dataset \cite{mvsec}. Following \cite{ramnet}, the evaluations are performed on the `outdoor\_day1', `outdoor\_night1', `outdoor\_night2', and `outdoor\_night3' sequences.

\paragraph{Results.} The comparisons of our methods and state-of-the-art methods with and without a cutoff threshold at 30 meters are given in \cref{tab:mvsecdepth} and \cref{tab:mvsecdepth_cutoff}, respectively.
Though the previous best method HMNet \cite{hmnet} performs supervised pre-training using ground-truth depth before fine-tuning on the MVSEC dataset, all our methods outperform it. For example, in \cref{tab:mvsecdepth_cutoff}, the averaged root mean squared error of HMNet is 7.534, while the errors of our methods with ResNet50, ViT-S/16, and Swin-T/7 backbones are  6.805, 6.957, and 6.870, respectively.

Sample prediction results of our method on the semantic segmentation, optical flow estimation, and depth estimation tasks are provided in \cref{fig:seg_mainpaper}.

\begin{table*}[!t]
    \centering
    \caption{\small \it (a)-(c) Comparison of state-of-the-art methods pre-trained on the N-imageNet datasets with backbones of ResNet50, ViT-S/16, Swin-T/7.  
    (d)-(f) Comparison of state-of-the-art methods pre-trained on the E-TartanAir datasets with backbones of ResNet50, ViT-S/16, Swin-T/7. 
    }
    \small
    \vspace{-.5em}
    \subfloat[
    All methods are pre-trained using the ResNet50 backbone, on the N-ImageNet dataset.
    \label{tab:dataset_a}]{
    \begin{minipage}{0.315\linewidth}
    \begin{tabularx}{\linewidth}{lYY}
        \toprule
         Method  & mIOU$\uparrow$ & mACC$\uparrow$\\
         \midrule
         SelfPath  & 57.881 & 64.916 \\
         ESViT & 57.796 & 64.928 \\
         ECDP  & 59.155 & 67.534\\
         Ours & \textbf{60.243} & \textbf{69.195}\\
        \bottomrule 
    \end{tabularx}
    \end{minipage}}
    \hspace{0.35em}
    \subfloat[All methods are pre-trained using the ViT-S/16 backbone, on the N-ImageNet dataset.
    \label{tab:dataset_b}]{
    \begin{minipage}{0.315\linewidth}
    \begin{tabularx}{\linewidth}{lYY}
        \toprule
         Method  & mIOU$\uparrow$ & mACC$\uparrow$\\
         \midrule
         SelfPatch  & 50.442 & 58.452 \\
         ESViT &  51.011 & 58.902 \\
         ECDP  &  52.517 & 60.553\\
         Ours & \textbf{54.897} & \textbf{62.527}\\
        \bottomrule 
    \end{tabularx}
    \end{minipage}}
    \hspace{0.35em}
    \subfloat[All methods are pre-trained using the Swin-T/7 backbone, on the N-ImageNet dataset.
    \label{tab:dataset_c}]{
    \begin{minipage}{0.315\linewidth}
    \begin{tabularx}{\linewidth}{lYY}
        \toprule
         Method  & mIOU$\uparrow$ & mACC$\uparrow$\\
         \midrule
         SelfPatch  & 52.997 & 59.928 \\
         ESViT & 53.051 & 60.094 \\
         ECDP  & 55.842 & 63.548 \\
         Ours & \textbf{56.654} & \textbf{65.250} \\
        \bottomrule 
    \end{tabularx}
    \end{minipage}}
    \\
    \subfloat[
    All methods are pre-trained using the ResNet50 backbone, on the E-TartanAir dataset.
    \label{tab:dataset_d}]{
    \begin{minipage}{0.315\linewidth}
    \begin{tabularx}{\linewidth}{lYY}
            \toprule
             Method  & mIOU$\uparrow$ & mACC$\uparrow$\\
             \midrule
             SelfPatch  &  58.365 & 65.180 \\
             ESViT  &  59.058 & 65.879 \\
             ECDP & 59.572 & 68.317\\
             Ours & \textbf{60.641} & \textbf{69.502} \\
            \bottomrule 
        \end{tabularx}
    \end{minipage}}
    \hspace{0.35em}
    \subfloat[
    All methods are pre-trained using the ViT-S/16 backbone, on the E-TartanAir dataset.
    \label{tab:dataset_e}]{
    \begin{minipage}{0.315\linewidth}
    \begin{tabularx}{\linewidth}{lYY}
            \toprule
             Method  & mIOU$\uparrow$ & mACC$\uparrow$\\
             \midrule
             SelfPatch & 52.347 & 59.947\\
             ESViT   & 51.945 &  60.470 \\
             ECDP  & 53.229 & 61.712 \\
             Ours  & \textbf{55.729} & \textbf{64.771}\\
            \bottomrule 
        \end{tabularx}
    \end{minipage}}
    \hspace{0.35em}
    \subfloat[
    All methods are pre-trained using the Swin-T/7 backbone, on the E-TartanAir dataset. 
    \label{tab:dataset_f}]{
    \begin{minipage}{0.315\linewidth}
    \begin{tabularx}{\linewidth}{lYY}
        \toprule
         Method  & mIOU$\uparrow$ & mACC$\uparrow$\\
         \midrule
         SelfPatch & 57.243 & 66.070\\
         ESViT   & 58.593 & 65.746 \\
         ECDP & 56.568 &  64.234\\
         Ours  & \textbf{61.250} & \textbf{69.620}\\
        \bottomrule 
    \end{tabularx}
    \end{minipage}}
    \label{tab:dataset}
    \vspace{-1.8em}
\end{table*}

\subsection{Discussions}
\label{sec:abl}
We perform ablations on the DSEC semantic segmentation dataset \cite{dsec,ess} to study our model components. 
We set the pre-training backbone and dataset to the Swin-T/7 and E-TartanAir dataset, except where otherwise indicated.

\begin{wrapfigure}{r}{7.5cm}
    \centering
    \includegraphics{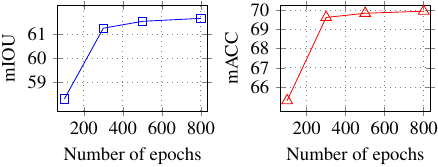}
    \caption{\small \it Comparison of the number of pre-training epochs.}
    \label{fig:pretraingepoch}
    \vspace{-.5cm}
\end{wrapfigure}
\paragraph{Pre-training datasets and backbones.} 
We investigate the impact of different pre-training datasets and backbones. We pre-train state-of-the-art methods with backbones of ResNet50, ViT-S/16, and Swin-T/7 respectively on the N-ImageNet and E-TartanAir datasets. The results are given in \cref{tab:dataset}. We have the following observations:
i) Our proposed E-TartanAir dataset outperforms the N-ImageNet dataset. The performance of each method gets improved, when changing the pretraining dataset from N-ImageNet to E-TartanAir; 
ii) Our method consistently achieves the best performance under each setting, and the best performance of our method is achieved with the Swin-T/7 backbone and E-TartanAir dataset. Furthermore, our method gains performance improvements by at least 0.812\% mIOU and 1.185\% mACC scores over the state-of-the-art methods, with the same pre-training backbone and dataset. Note that ESViT exhibits numerical instability during pre-training, and the restart strategy from \cite{opt} is used for pre-training.

\paragraph{Pre-training epochs.} We explore the impact of pre-training epochs, ranging from 100 to 800, and the results are presented in \cref{fig:pretraingepoch}. Limited performance improvements in mIOU and mACC scores are observed after 300 epochs, prompting us to set the pre-training epoch number to 300. 

\begin{wraptable}{r}{7.5cm}
    \vspace{-1em}
    \centering
    \caption{\small \it Comparison of the performance of networks trained with and without using the proposed context-level similarity loss $\mathcal{L}_{\text{context}}$. Using $\mathcal{L}_{\text{context}}$ consistently improves accuracies. `Pre. Dataset' and `\#Param'  respectively denote the pre-training dataset and the number of backbone parameters.  
    }
    \vspace{.5em}
    \small
    \begin{tabularx}{\linewidth}{lcYYY}
        \toprule
         \makecell[l]{Pre.\\Dataset} & Backbone & \#Param & mIOU$\uparrow$ & mACC$\uparrow$\\
         \midrule
         \multicolumn{5}{l}{\gc{$\mathcal{L}_{\text{patch}} + \mathcal{L}_{\text{image}}$.}} \\ 
           N-ImageNet  & ResNet50 & 23M & 58.308 & 65.597 \\
          {N-ImageNet} & {ViT-S/16} & {21M} & {53.706} & {61.328} \\
          {N-ImageNet} & {Swin-T/7} & {28M} & {54.905} & {63.271} \\
          E-TartanAir & ResNet50 & 23M & 58.687 & 66.171 \\
          {E-TartanAir} & {ViT-S/16} & {21M} & {54.193} & {61.711} \\
         {E-TartanAir} & {Swin-T/7} & {28M} & {55.556} & {63.486} \\
         \midrule
         \multicolumn{5}{l}{\gc{$\mathcal{L}_{\text{patch}} + \mathcal{L}_{\text{context}} + \mathcal{L}_{\text{image}}$.}} \\
         N-ImageNet  & ResNet50 & 23M & 60.243 & 69.195 \\
         N-ImageNet & ViT-S/16 & 21M & 54.897 & 62.527\\
         N-ImageNet & Swin-T/7 & 28M & 56.654 & 65.250\\
         E-TartanAir & ResNet50 & 23M  & 60.641 & 69.502  \\
         E-TartanAir & ViT-S/16 & 21M & 55.729 & 64.771\\
         E-TartanAir & Swin-T/7 & 28M & \textbf{61.250} & \textbf{69.620}  \\
        \bottomrule 
    \end{tabularx}
    \label{tab:vit}
    \vspace{-.5cm}
\end{wraptable}
\paragraph{Context-level similarity.}  
To check the effectiveness of our context-level similarity loss, we pre-train several networks without using it, varying our backbone network and pre-training dataset. Results in \cref{tab:vit} reveal that a network pre-trained with $\mathcal{L}_{\text{context}}$ consistently outperforms its counterpart pre-trained without using $\mathcal{L}_{\text{context}}$. For example, for networks pre-trained on the E-tartanAir dataset with the Swin-T/7 backbone, without using $\mathcal{L}_{\text{context}}$ in pre-training, the mIOU/mACC scores are 55.556\%/63.486\%, which are lower than our best scores of 61.250\%/69.620\%. This justifies the effectiveness of the proposed context-level similarity loss. 

\begin{figure*}[!t]
    \centering
    \bgroup
    \setlength{\tabcolsep}{3.5pt}
    \def\arraystretch{1.5}
        \begin{tabular}{cccccc}
        \includegraphics[width=0.185\linewidth]{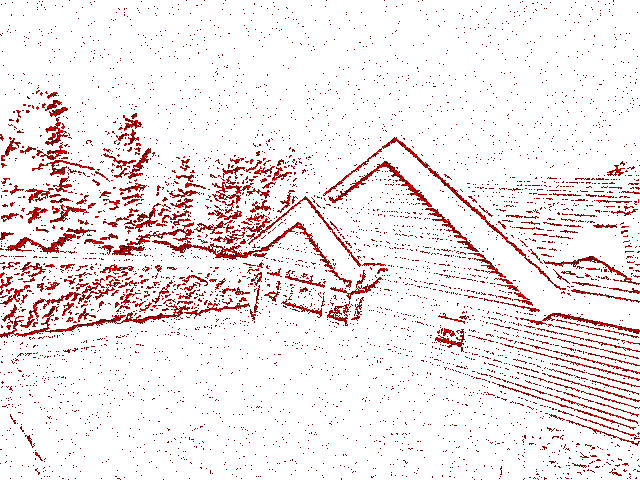} &
        \includegraphics[width=0.185\linewidth]{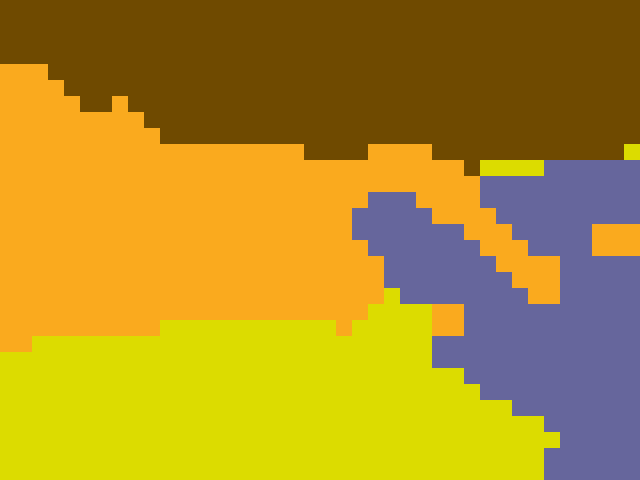} &
        \includegraphics[width=0.185\linewidth]{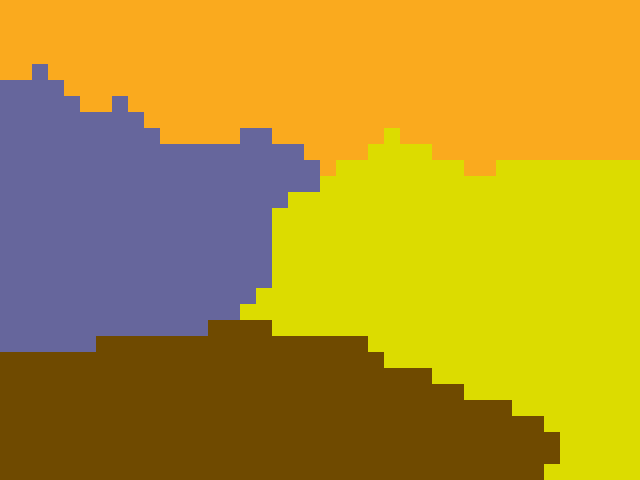} &
        \includegraphics[width=0.185\linewidth]{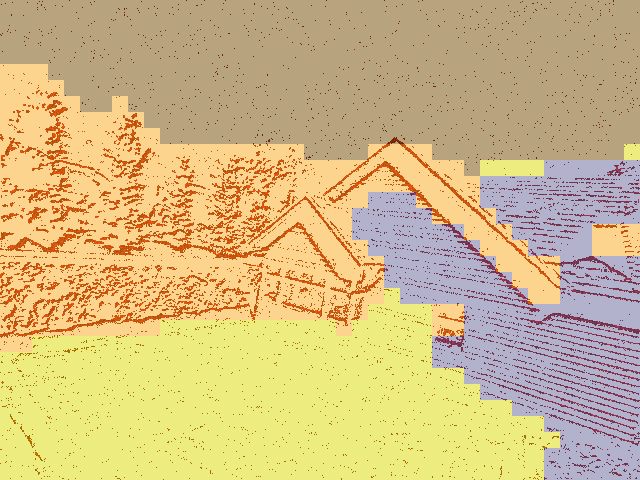} &
        \includegraphics[width=0.185\linewidth]{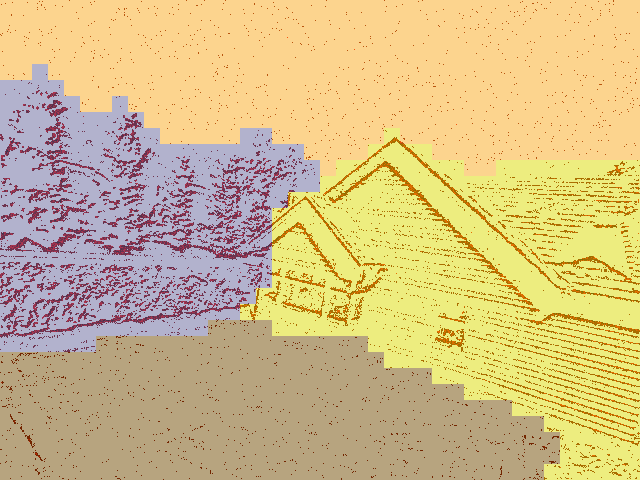}  \\
       \includegraphics[width=0.185\linewidth]{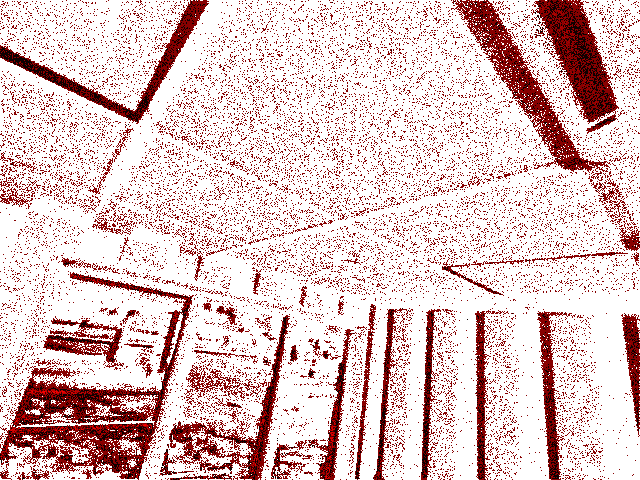} &
        \includegraphics[width=0.185\linewidth]{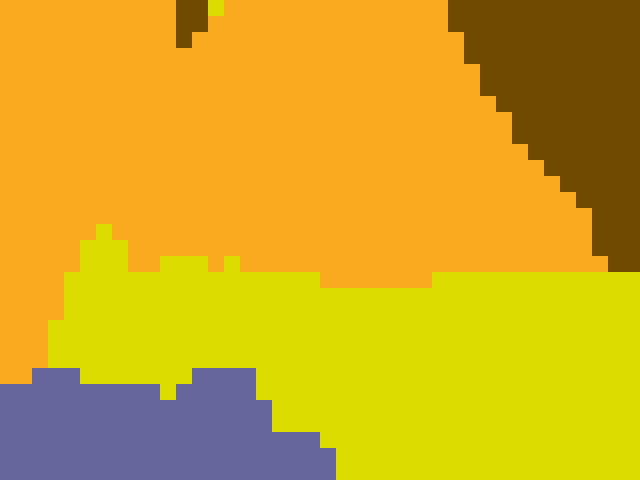} &
        \includegraphics[width=0.185\linewidth]{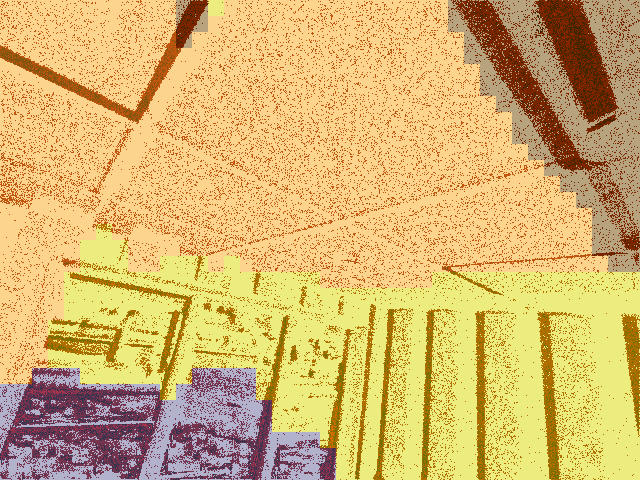} &
        \includegraphics[width=0.185\linewidth]{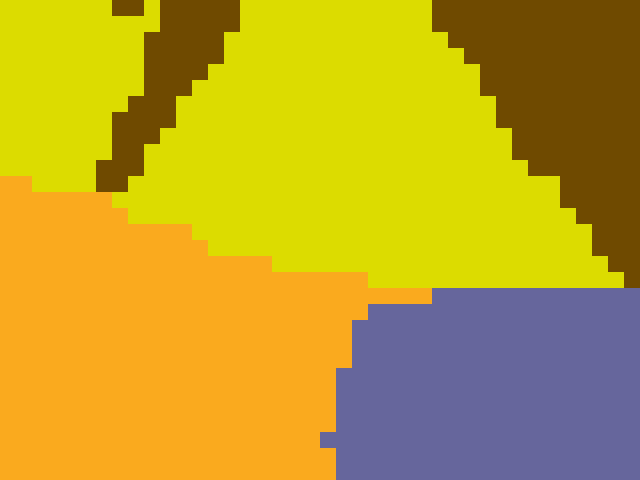} &
        \includegraphics[width=0.185\linewidth]{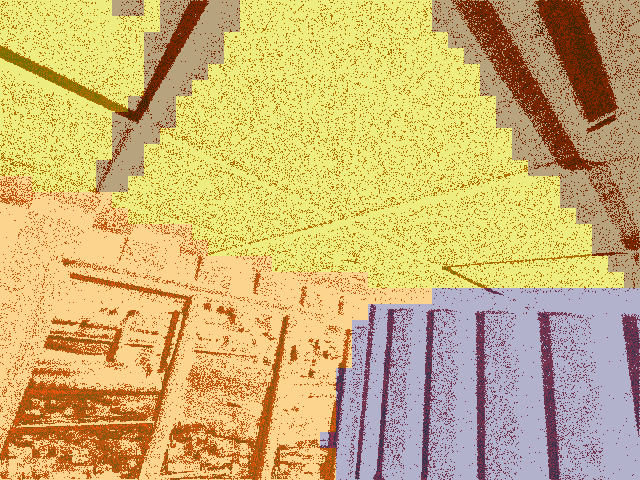}  \\
      (a) & (b) & (c) &  (d) & (e) \\
    \end{tabular}
    \egroup
    \vspace{-.5em}
    \caption{\small \it Sample results of patches belonging to different contexts on the E-TartanAir dataset. (a): input event images. (b): mined context labels (without enforcing the context-level similarity). (c): mined context labels (enforcing the context-level similarity). (d) and (e): blends of the event image with context labels from (b) and (c) for visualization purposes, respectively.
    }
    \label{fig:cluster}
\end{figure*}

Sample results of patches belonging to different contexts are given in \cref{fig:cluster}. For example, in the 1$^{\text{st}}$ row, our method successfully mines contexts (tree, building, ground, and sky) in an event image, and groups patches with the same semantics.

\begin{wrapfigure}{r}{7.5cm}
    \centering
    \vspace{-.75cm}
    \includegraphics{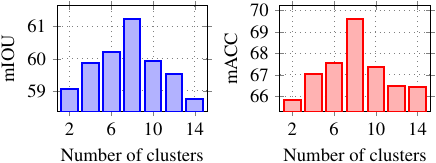}
    \caption{\small \it 
    Comparison of the number of contexts.
    }
    \label{fig:numberofcluster}
    \vspace{-.5cm}
\end{wrapfigure}

\paragraph{Number of contexts.} Our model utilizes $K$ context embeddings by aggregating patch features.
To study the impact of contexts, we train our model with a different number of contexts. Results in \cref{fig:numberofcluster} indicate that the best performance is achieved with $8$ contexts.
Increasing the number of contexts
results in inferior performance. Due to the sparsity of event camera data, for large context numbers, many contexts aggregate features from event patches with little to no events. This results in noisy context embeddings, interferes with the training process, and hinders the network from learning discriminative event features.

\paragraph{Generalization ability of context-level similarity.} To further demonstrate the effectiveness of the proposed  $\mathcal{L}_{\text{context}}$, we add $\mathcal{L}_{\text{context}}$ to the objective function of the state-of-the-art event data pre-training method, ECDP \cite{eventpretraing}. The mIOU/mAcc scores of ECDP are increased by a large margin, having improvements from 52.517\%/60.553\% to 53.826\%/61.008\%. The improvements validate the generalization ability of $\mathcal{L}_{\text{context}}$.

\paragraph{Limitation.} Although our self-supervised pre-trained network has achieved state-of-the-art performance across various dense prediction tasks, it necessitates task-specific fine-tuning to refine pre-trained network weights. However, we believe that our self-supervised learning exploration helps to learn task-agnostic pre-trained representations.

\section{Conclusion and Broader Impact} We present a neural network trained for dense prediction tasks using an event camera. Our self-supervised learning method enforces three levels of similarity constraints: patch-level, context-level, and image-level. Our key insight is {enforcing context similarity} from event patch embeddings to pre-train our model. The proposed context-level similarity effectively addresses the sparsity problem of event data, resulting in state-of-the-art performance on semantic segmentation, optical flow, and depth estimation benchmarks. We believe that our dense pre-training techniques deserve a position in highly accurate event-based dense predictions.

\paragraph{Broader Impact.} By aligning event data with paired RGB frames, our pre-training framework is promising to be extended to an event-vision-language foundation model. We hope it inspires future work.

%\clearpage  % TODO FINAL: This \clearpage needs to be removed from both review and camera-ready versions.

\section*{Acknowledgements}
Liyuan Pan's work was supported in part by the Beijing Institute of Technology Research Fund Program for Young Scholars, BIT Special-Zone, and National Natural Science Foundation of China 62302045.

\bibliographystyle{splncs04}
\bibliography{main}

\begin{thebibliography}{10}
\providecommand{\url}[1]{\texttt{#1}}
\providecommand{\urlprefix}{URL }
\providecommand{\doi}[1]{https://doi.org/#1}

\bibitem{evsegnet}
Alonso, I., Murillo, A.C.: Ev-segnet: Semantic segmentation for event-based cameras. In: {IEEE} Conference on Computer Vision and Pattern Recognition Workshops, {CVPR} Workshops 2019, Long Beach, CA, USA, June 16-20, 2019. pp. 1624--1633. Computer Vision Foundation / {IEEE} (2019). \doi{10.1109/CVPRW.2019.00205}, \url{http://openaccess.thecvf.com/content\_CVPRW\_2019/html/EventVision/Alonso\_EV-SegNet\_Semantic\_Segmentation\_for\_Event-Based\_Cameras\_CVPRW\_2019\_paper.html}

\bibitem{pointcl}
Bai, Y., Chen, X., Kirillov, A., Yuille, A.L., Berg, A.C.: Point-level region contrast for object detection pre-training. In: {IEEE/CVF} Conference on Computer Vision and Pattern Recognition, {CVPR} 2022, New Orleans, LA, USA, June 18-24, 2022. pp. 16040--16049. {IEEE} (2022). \doi{10.1109/CVPR52688.2022.01559}, \url{https://doi.org/10.1109/CVPR52688.2022.01559}

\bibitem{beit}
Bao, H., Dong, L., Piao, S., Wei, F.: Beit: {BERT} pre-training of image transformers. In: The Tenth International Conference on Learning Representations, {ICLR} 2022, Virtual Event, April 25-29, 2022. OpenReview.net (2022), \url{https://openreview.net/forum?id=p-BhZSz59o4}

\bibitem{ddd17}
Binas, J., Neil, D., Liu, S., Delbr{\"{u}}ck, T.: {DDD17:} end-to-end {DAVIS} driving dataset. CoRR  \textbf{abs/1711.01458} (2017), \url{http://arxiv.org/abs/1711.01458}

\bibitem{swav}
Caron, M., Misra, I., Mairal, J., Goyal, P., Bojanowski, P., Joulin, A.: Unsupervised learning of visual features by contrasting cluster assignments. In: Larochelle, H., Ranzato, M., Hadsell, R., Balcan, M., Lin, H. (eds.) Advances in Neural Information Processing Systems 33: Annual Conference on Neural Information Processing Systems 2020, NeurIPS 2020, December 6-12, 2020, virtual (2020), \url{https://proceedings.neurips.cc/paper/2020/hash/70feb62b69f16e0238f741fab228fec2-Abstract.html}

\bibitem{dino}
Caron, M., Touvron, H., Misra, I., J{\'{e}}gou, H., Mairal, J., Bojanowski, P., Joulin, A.: Emerging properties in self-supervised vision transformers. In: 2021 {IEEE/CVF} International Conference on Computer Vision, {ICCV} 2021, Montreal, QC, Canada, October 10-17, 2021. pp. 9630--9640. {IEEE} (2021). \doi{10.1109/ICCV48922.2021.00951}, \url{https://doi.org/10.1109/ICCV48922.2021.00951}

\bibitem{simclr}
Chen, T., Kornblith, S., Norouzi, M., Hinton, G.E.: A simple framework for contrastive learning of visual representations. In: Proceedings of the 37th International Conference on Machine Learning, {ICML} 2020, 13-18 July 2020, Virtual Event. Proceedings of Machine Learning Research, vol.~119, pp. 1597--1607. {PMLR} (2020), \url{http://proceedings.mlr.press/v119/chen20j.html}

\bibitem{mocov2}
Chen, X., Fan, H., Girshick, R., He, K.: Improved baselines with momentum contrastive learning. arXiv preprint arXiv:2003.04297  (2020)

\bibitem{simsiam}
Chen, X., He, K.: Exploring simple siamese representation learning. In: {IEEE} Conference on Computer Vision and Pattern Recognition, {CVPR} 2021, virtual, June 19-25, 2021. pp. 15750--15758. Computer Vision Foundation / {IEEE} (2021). \doi{10.1109/CVPR46437.2021.01549}, \url{https://openaccess.thecvf.com/content/CVPR2021/html/Chen\_Exploring\_Simple\_Siamese\_Representation\_Learning\_CVPR\_2021\_paper.html}

\bibitem{mocov3}
Chen*, X., Xie*, S., He, K.: An empirical study of training self-supervised vision transformers. arXiv preprint arXiv:2104.02057  (2021)

\bibitem{CIFAR-10-DVS}
Cheng, W., Luo, H., Yang, W., Yu, L., Li, W.: Structure-aware network for lane marker extraction with dynamic vision sensor. CoRR  \textbf{abs/2008.06204} (2020), \url{https://arxiv.org/abs/2008.06204}

\bibitem{scalinglaw}
Cherti, M., Beaumont, R., Wightman, R., Wortsman, M., Ilharco, G., Gordon, C., Schuhmann, C., Schmidt, L., Jitsev, J.: Reproducible scaling laws for contrastive language-image learning. In: {IEEE/CVF} Conference on Computer Vision and Pattern Recognition, {CVPR} 2023, Vancouver, BC, Canada, June 17-24, 2023. pp. 2818--2829. {IEEE} (2023). \doi{10.1109/CVPR52729.2023.00276}, \url{https://doi.org/10.1109/CVPR52729.2023.00276}

\bibitem{onevflow}
Cuadrado, J., Rancon, U., Cottereau, B., Barranco, F., Masquelier, T.: Optical flow estimation from event-based cameras and spiking neural networks. Frontiers in Neuroscience  \textbf{17} (05 2023). \doi{10.3389/fnins.2023.1160034}

\bibitem{imagenet}
Deng, J., Dong, W., Socher, R., Li, L., Li, K., Fei{-}Fei, L.: Imagenet: {A} large-scale hierarchical image database. In: 2009 {IEEE} Computer Society Conference on Computer Vision and Pattern Recognition {(CVPR} 2009), 20-25 June 2009, Miami, Florida, {USA}. pp. 248--255. {IEEE} Computer Society (2009). \doi{10.1109/CVPR.2009.5206848}, \url{https://doi.org/10.1109/CVPR.2009.5206848}

\bibitem{vit}
Dosovitskiy, A., Beyer, L., Kolesnikov, A., Weissenborn, D., Zhai, X., Unterthiner, T., Dehghani, M., Minderer, M., Heigold, G., Gelly, S., Uszkoreit, J., Houlsby, N.: An image is worth 16x16 words: Transformers for image recognition at scale. In: 9th International Conference on Learning Representations, {ICLR} 2021, Virtual Event, Austria, May 3-7, 2021. OpenReview.net (2021), \url{https://openreview.net/forum?id=YicbFdNTTy}

\bibitem{vqgan}
Esser, P., Rombach, R., Ommer, B.: Taming transformers for high-resolution image synthesis. In: {IEEE} Conference on Computer Vision and Pattern Recognition, {CVPR} 2021, virtual, June 19-25, 2021. pp. 12873--12883. Computer Vision Foundation / {IEEE} (2021). \doi{10.1109/CVPR46437.2021.01268}, \url{https://openaccess.thecvf.com/content/CVPR2021/html/Esser\_Taming\_Transformers\_for\_High-Resolution\_Image\_Synthesis\_CVPR\_2021\_paper.html}

\bibitem{eva}
Fang, Y., Wang, W., Xie, B., Sun, Q., Wu, L., Wang, X., Huang, T., Wang, X., Cao, Y.: {EVA:} exploring the limits of masked visual representation learning at scale. In: {IEEE/CVF} Conference on Computer Vision and Pattern Recognition, {CVPR} 2023, Vancouver, BC, Canada, June 17-24, 2023. pp. 19358--19369. {IEEE} (2023). \doi{10.1109/CVPR52729.2023.01855}, \url{https://doi.org/10.1109/CVPR52729.2023.01855}

\bibitem{eventsurvey}
Gallego, G., Delbr{\"{u}}ck, T., Orchard, G., Bartolozzi, C., Taba, B., Censi, A., Leutenegger, S., Davison, A.J., Conradt, J., Daniilidis, K., Scaramuzza, D.: Event-based vision: {A} survey. {IEEE} Trans. Pattern Anal. Mach. Intell.  \textbf{44}(1),  154--180 (2022). \doi{10.1109/TPAMI.2020.3008413}, \url{https://doi.org/10.1109/TPAMI.2020.3008413}

\bibitem{ramnet}
Gehrig, D., R{\"{u}}egg, M., Gehrig, M., Hidalgo{-}Carri{\'{o}}, J., Scaramuzza, D.: Combining events and frames using recurrent asynchronous multimodal networks for monocular depth prediction. {IEEE} Robotics Autom. Lett.  \textbf{6}(2),  2822--2829 (2021). \doi{10.1109/LRA.2021.3060707}, \url{https://doi.org/10.1109/LRA.2021.3060707}

\bibitem{dsec}
Gehrig, M., Aarents, W., Gehrig, D., Scaramuzza, D.: {DSEC:} {A} stereo event camera dataset for driving scenarios. {IEEE} Robotics Autom. Lett.  \textbf{6}(3),  4947--4954 (2021). \doi{10.1109/LRA.2021.3068942}, \url{https://doi.org/10.1109/LRA.2021.3068942}

\bibitem{eraft}
Gehrig, M., Millh{\"{a}}usler, M., Gehrig, D., Scaramuzza, D.: {E-RAFT:} dense optical flow from event cameras. In: International Conference on 3D Vision, 3DV 2021, London, United Kingdom, December 1-3, 2021. pp. 197--206. {IEEE} (2021). \doi{10.1109/3DV53792.2021.00030}, \url{https://doi.org/10.1109/3DV53792.2021.00030}

\bibitem{byol}
Grill, J., Strub, F., Altch{\'{e}}, F., Tallec, C., Richemond, P.H., Buchatskaya, E., Doersch, C., Pires, B.{\'{A}}., Guo, Z., Azar, M.G., Piot, B., Kavukcuoglu, K., Munos, R., Valko, M.: Bootstrap your own latent - {A} new approach to self-supervised learning. In: Larochelle, H., Ranzato, M., Hadsell, R., Balcan, M., Lin, H. (eds.) Advances in Neural Information Processing Systems 33: Annual Conference on Neural Information Processing Systems 2020, NeurIPS 2020, December 6-12, 2020, virtual (2020), \url{https://proceedings.neurips.cc/paper/2020/hash/f3ada80d5c4ee70142b17b8192b2958e-Abstract.html}

\bibitem{hmnet}
Hamaguchi, R., Furukawa, Y., Onishi, M., Sakurada, K.: Hierarchical neural memory network for low latency event processing. In: {IEEE/CVF} Conference on Computer Vision and Pattern Recognition, {CVPR} 2023, Vancouver, BC, Canada, June 17-24, 2023. pp. 22867--22876. {IEEE} (2023). \doi{10.1109/CVPR52729.2023.02190}, \url{https://doi.org/10.1109/CVPR52729.2023.02190}

\bibitem{mae}
He, K., Chen, X., Xie, S., Li, Y., Doll{\'{a}}r, P., Girshick, R.B.: Masked autoencoders are scalable vision learners. In: {IEEE/CVF} Conference on Computer Vision and Pattern Recognition, {CVPR} 2022, New Orleans, LA, USA, June 18-24, 2022. pp. 15979--15988. {IEEE} (2022). \doi{10.1109/CVPR52688.2022.01553}, \url{https://doi.org/10.1109/CVPR52688.2022.01553}

\bibitem{mocov1}
He, K., Fan, H., Wu, Y., Xie, S., Girshick, R.B.: Momentum contrast for unsupervised visual representation learning. In: 2020 {IEEE/CVF} Conference on Computer Vision and Pattern Recognition, {CVPR} 2020, Seattle, WA, USA, June 13-19, 2020. pp. 9726--9735. Computer Vision Foundation / {IEEE} (2020). \doi{10.1109/CVPR42600.2020.00975}, \url{https://doi.org/10.1109/CVPR42600.2020.00975}

\bibitem{resnet}
He, K., Zhang, X., Ren, S., Sun, J.: Deep residual learning for image recognition. CoRR  \textbf{abs/1512.03385} (2015), \url{http://arxiv.org/abs/1512.03385}

\bibitem{nimagnet}
Kim, J., Bae, J., Park, G., Zhang, D., Kim, Y.M.: N-imagenet: Towards robust, fine-grained object recognition with event cameras. In: Proceedings of the IEEE/CVF International Conference on Computer Vision (ICCV). pp. 2146--2156 (October 2021)

\bibitem{esvit}
Li, C., Yang, J., Zhang, P., Gao, M., Xiao, B., Dai, X., Yuan, L., Gao, J.: Efficient self-supervised vision transformers for representation learning. In: The Tenth International Conference on Learning Representations, {ICLR} 2022, Virtual Event, April 25-29, 2022. OpenReview.net (2022), \url{https://openreview.net/forum?id=fVu3o-YUGQK}

\bibitem{cim}
Li, W., Xie, J., Loy, C.C.: Correlational image modeling for self-supervised visual pre-training. In: {IEEE/CVF} Conference on Computer Vision and Pattern Recognition, {CVPR} 2023, Vancouver, BC, Canada, June 17-24, 2023. pp. 15105--15115. {IEEE} (2023). \doi{10.1109/CVPR52729.2023.01450}, \url{https://doi.org/10.1109/CVPR52729.2023.01450}

\bibitem{eflowformer}
Li, Y., Huang, Z., Chen, S., Shi, X., Li, H., Bao, H., Cui, Z., Zhang, G.: Blinkflow: {A} dataset to push the limits of event-based optical flow estimation. CoRR  \textbf{abs/2303.07716} (2023). \doi{10.48550/arXiv.2303.07716}, \url{https://doi.org/10.48550/arXiv.2303.07716}

\bibitem{tma}
Liu, H., Chen, G., Qu, S., Zhang, Y., Li, Z., Knoll, A., Jiang, C.: Tma: Temporal motion aggregation for event-based optical flow. In: ICCV (2023)

\bibitem{kitti}
Menze, M., Heipke, C., Geiger, A.: Joint 3d estimation of vehicles and scene flow. ISPRS Annals of Photogrammetry, Remote Sensing and Spatial Information Sciences  \textbf{II-3/W5},  427--434 (08 2015). \doi{10.5194/isprsannals-II-3-W5-427-2015}

\bibitem{dinov2}
Oquab, M., Darcet, T., Moutakanni, T., Vo, H., Szafraniec, M., Khalidov, V., Fernandez, P., Haziza, D., Massa, F., El{-}Nouby, A., Assran, M., Ballas, N., Galuba, W., Howes, R., Huang, P., Li, S., Misra, I., Rabbat, M.G., Sharma, V., Synnaeve, G., Xu, H., J{\'{e}}gou, H., Mairal, J., Labatut, P., Joulin, A., Bojanowski, P.: Dinov2: Learning robust visual features without supervision. CoRR  \textbf{abs/2304.07193} (2023). \doi{10.48550/arXiv.2304.07193}, \url{https://doi.org/10.48550/arXiv.2304.07193}

\bibitem{ncaltech}
Orchard, G., Jayawant, A., Cohen, G., Thakor, N.V.: Converting static image datasets to spiking neuromorphic datasets using saccades. CoRR  \textbf{abs/1507.07629} (2015), \url{http://arxiv.org/abs/1507.07629}

\bibitem{beitv2}
Peng, Z., Dong, L., Bao, H., Ye, Q., Wei, F.: Beit v2: Masked image modeling with vector-quantized visual tokenizers. CoRR  \textbf{abs/2208.06366} (2022). \doi{10.48550/arXiv.2208.06366}, \url{https://doi.org/10.48550/arXiv.2208.06366}

\bibitem{clip}
Radford, A., Kim, J.W., Hallacy, C., Ramesh, A., Goh, G., Agarwal, S., Sastry, G., Askell, A., Mishkin, P., Clark, J., Krueger, G., Sutskever, I.: Learning transferable visual models from natural language supervision. In: Meila, M., Zhang, T. (eds.) Proceedings of the 38th International Conference on Machine Learning, {ICML} 2021, 18-24 July 2021, Virtual Event. Proceedings of Machine Learning Research, vol.~139, pp. 8748--8763. {PMLR} (2021), \url{http://proceedings.mlr.press/v139/radford21a.html}

\bibitem{multicm}
Shiba, S., Aoki, Y., Gallego, G.: Secrets of event-based optical flow. In: Avidan, S., Brostow, G.J., Ciss{\'{e}}, M., Farinella, G.M., Hassner, T. (eds.) Computer Vision - {ECCV} 2022 - 17th European Conference, Tel Aviv, Israel, October 23-27, 2022, Proceedings, Part {XVIII}. Lecture Notes in Computer Science, vol. 13678, pp. 628--645. Springer (2022). \doi{10.1007/978-3-031-19797-0\_36}, \url{https://doi.org/10.1007/978-3-031-19797-0\_36}

\bibitem{ncars}
Sironi, A., Brambilla, M., Bourdis, N., Lagorce, X., Benosman, R.: {HATS:} histograms of averaged time surfaces for robust event-based object classification. In: 2018 {IEEE} Conference on Computer Vision and Pattern Recognition, {CVPR} 2018, Salt Lake City, UT, USA, June 18-22, 2018. pp. 1731--1740. Computer Vision Foundation / {IEEE} Computer Society (2018). \doi{10.1109/CVPR.2018.00186}, \url{http://openaccess.thecvf.com/content\_cvpr\_2018/html/Sironi\_HATS\_Histograms\_of\_CVPR\_2018\_paper.html}

\bibitem{ess}
Sun, Z., Messikommer, N., Gehrig, D., Scaramuzza, D.: {ESS:} learning event-based semantic segmentation from still images. In: Avidan, S., Brostow, G.J., Ciss{\'{e}}, M., Farinella, G.M., Hassner, T. (eds.) Computer Vision - {ECCV} 2022 - 17th European Conference, Tel Aviv, Israel, October 23-27, 2022, Proceedings, Part {XXXIV}. Lecture Notes in Computer Science, vol. 13694, pp. 341--357. Springer (2022). \doi{10.1007/978-3-031-19830-4\_20}, \url{https://doi.org/10.1007/978-3-031-19830-4\_20}

\bibitem{dceiflow}
Wan, Z., Dai, Y., Mao, Y.: Learning dense and continuous optical flow from an event camera. {IEEE} Trans. Image Process.  \textbf{31},  7237--7251 (2022). \doi{10.1109/TIP.2022.3220938}, \url{https://doi.org/10.1109/TIP.2022.3220938}

\bibitem{tartanair}
Wang, W., Zhu, D., Wang, X., Hu, Y., Qiu, Y., Wang, C., Hu, Y., Kapoor, A., Scherer, S.A.: Tartanair: {A} dataset to push the limits of visual {SLAM}. In: {IEEE/RSJ} International Conference on Intelligent Robots and Systems, {IROS} 2020, Las Vegas, NV, USA, October 24, 2020 - January 24, 2021. pp. 4909--4916. {IEEE} (2020). \doi{10.1109/IROS45743.2020.9341801}, \url{https://doi.org/10.1109/IROS45743.2020.9341801}

\bibitem{densecl}
Wang, X., Zhang, R., Shen, C., Kong, T., Li, L.: Dense contrastive learning for self-supervised visual pre-training. In: {IEEE} Conference on Computer Vision and Pattern Recognition, {CVPR} 2021, virtual, June 19-25, 2021. pp. 3024--3033. Computer Vision Foundation / {IEEE} (2021). \doi{10.1109/CVPR46437.2021.00304}, \url{https://openaccess.thecvf.com/content/CVPR2021/html/Wang\_Dense\_Contrastive\_Learning\_for\_Self-Supervised\_Visual\_Pre-Training\_CVPR\_2021\_paper.html}

\bibitem{slam}
Weikersdorfer, D., Adrian, D.B., Cremers, D., Conradt, J.: Event-based 3d {SLAM} with a depth-augmented dynamic vision sensor. In: 2014 {IEEE} International Conference on Robotics and Automation, {ICRA} 2014, Hong Kong, China, May 31 - June 7, 2014. pp. 359--364. {IEEE} (2014). \doi{10.1109/ICRA.2014.6906882}, \url{https://doi.org/10.1109/ICRA.2014.6906882}

\bibitem{idnet}
Wu, Y., Paredes-Vall\'es, F., de~Croon, G.C.H.E.: Lightweight event-based optical flow estimation via iterative deblurring. In: Proceedings of IEEE International Conference on Robotics and Automation (ICRA'24) (May 2024), to Appear

\bibitem{instancediscrmination}
Wu, Z., Xiong, Y., Yu, S.X., Lin, D.: Unsupervised feature learning via non-parametric instance discrimination. In: 2018 {IEEE} Conference on Computer Vision and Pattern Recognition, {CVPR} 2018, Salt Lake City, UT, USA, June 18-22, 2018. pp. 3733--3742. Computer Vision Foundation / {IEEE} Computer Society (2018). \doi{10.1109/CVPR.2018.00393}, \url{http://openaccess.thecvf.com/content\_cvpr\_2018/html/Wu\_Unsupervised\_Feature\_Learning\_CVPR\_2018\_paper.html}

\bibitem{simmim}
Xie, Z., Zhang, Z., Cao, Y., Lin, Y., Bao, J., Yao, Z., Dai, Q., Hu, H.: Simmim: a simple framework for masked image modeling. In: {IEEE/CVF} Conference on Computer Vision and Pattern Recognition, {CVPR} 2022, New Orleans, LA, USA, June 18-24, 2022. pp. 9643--9653. {IEEE} (2022). \doi{10.1109/CVPR52688.2022.00943}, \url{https://doi.org/10.1109/CVPR52688.2022.00943}

\bibitem{eventpretraing}
Yang, Y., Pan, L., Liu, L.: Event camera data pre-training. In: Proceedings of the IEEE/CVF International Conference on Computer Vision (ICCV). pp. 10699--10709 (October 2023)

\bibitem{selfpatch}
Yun, S., Lee, H., Kim, J., Shin, J.: Patch-level representation learning for self-supervised vision transformers. In: {IEEE/CVF} Conference on Computer Vision and Pattern Recognition, {CVPR} 2022, New Orleans, LA, USA, June 18-24, 2022. pp. 8344--8353. {IEEE} (2022). \doi{10.1109/CVPR52688.2022.00817}, \url{https://doi.org/10.1109/CVPR52688.2022.00817}

\bibitem{event_image_association}
Zhang, D., Ding, Q., Duan, P., Zhou, C., Shi, B.: Data association between event streams and intensity frames under diverse baselines. In: Avidan, S., Brostow, G.J., Ciss{\'{e}}, M., Farinella, G.M., Hassner, T. (eds.) Computer Vision - {ECCV} 2022 - 17th European Conference, Tel Aviv, Israel, October 23-27, 2022, Proceedings, Part {VII}. Lecture Notes in Computer Science, vol. 13667, pp. 72--90. Springer (2022). \doi{10.1007/978-3-031-20071-7\_5}, \url{https://doi.org/10.1007/978-3-031-20071-7\_5}

\bibitem{opt}
Zhang, S., Roller, S., Goyal, N., Artetxe, M., Chen, M., Chen, S., Dewan, C., Diab, M.T., Li, X., Lin, X.V., Mihaylov, T., Ott, M., Shleifer, S., Shuster, K., Simig, D., Koura, P.S., Sridhar, A., Wang, T., Zettlemoyer, L.: {OPT:} open pre-trained transformer language models. CoRR  \textbf{abs/2205.01068} (2022). \doi{10.48550/ARXIV.2205.01068}, \url{https://doi.org/10.48550/arXiv.2205.01068}

\bibitem{eclip}
Zhou, J., Zheng, X., Lyu, Y., Wang, L.: {E-CLIP:} towards label-efficient event-based open-world understanding by {CLIP}. CoRR  \textbf{abs/2308.03135} (2023). \doi{10.48550/arXiv.2308.03135}, \url{https://doi.org/10.48550/arXiv.2308.03135}

\bibitem{ibot}
Zhou, J., Wei, C., Wang, H., Shen, W., Xie, C., Yuille, A.L., Kong, T.: Image {BERT} pre-training with online tokenizer. In: The Tenth International Conference on Learning Representations, {ICLR} 2022, Virtual Event, April 25-29, 2022. OpenReview.net (2022), \url{https://openreview.net/forum?id=ydopy-e6Dg}

\bibitem{mvsec}
Zhu, A.Z., Thakur, D., {\"{O}}zaslan, T., Pfrommer, B., Kumar, V., Daniilidis, K.: The multi vehicle stereo event camera dataset: An event camera dataset for 3d perception. CoRR  \textbf{abs/1801.10202} (2018), \url{http://arxiv.org/abs/1801.10202}

\bibitem{voxelgrid}
Zhu, A.Z., Yuan, L., Chaney, K., Daniilidis, K.: Unsupervised event-based learning of optical flow, depth, and egomotion. CoRR  \textbf{abs/1812.08156} (2018), \url{http://arxiv.org/abs/1812.08156}

\end{thebibliography}
\end{document}